\documentclass[10pt,twocolumn,letterpaper]{article}

\usepackage[pagenumbers]{cvpr} %

\usepackage{microtype}

\usepackage{amssymb}
\usepackage{pifont}
\newcommand{\cmark}{\ding{51}}
\newcommand{\xmark}{\ding{55}}
\usepackage{graphicx}
\usepackage{booktabs}
\usepackage{multirow}
\usepackage{array}

\definecolor{cSpeaker}{HTML}{2F6DB5} 
\definecolor{cMotion}{HTML}{A8801A} 
\definecolor{cRelation}{HTML}{B0479B} 
\definecolor{cLang}{HTML}{C8643B} 
\definecolor{cAgg}{HTML}{8A4FB0} 
\definecolor{cCandidate}{HTML}{D62728}
\definecolor{cTarget}{HTML}{D62828}

\usepackage{xcolor}
\usepackage[table]{xcolor}

\definecolor{okbg}{RGB}{226,240,217}
\definecolor{nobg}{RGB}{252,228,228}
\definecolor{ptbg}{RGB}{253,235,208}
\definecolor{okfg}{RGB}{46,125,50}
\definecolor{nofg}{RGB}{211,47,47}
\definecolor{ptfg}{RGB}{230,100,20}

\newcommand{\yes}{\cellcolor{okbg}\textcolor{okfg}{\ding{51}}}
\newcommand{\no}{\cellcolor{nobg}\textcolor{nofg}{\ding{55}}}
\newcommand{\repl}{\cellcolor{ptbg}\textcolor{ptfg}{(\ding{51})}}
\newcommand{\replmark}{\textcolor{ptfg}{(\ding{51})}}
\newcommand{\nomark}{\textcolor{nofg}{\ding{55}}}

\renewcommand{\cmark}{\textcolor{okfg}{\ding{51}}}
\renewcommand{\xmark}{\textcolor{nofg}{\ding{55}}}

\newif\ifshowchanges
\ifshowchanges
  \newcommand{\new}[1]{{\color{blue}#1}}
\else
  \newcommand{\new}[1]{#1}
\fi

\newcommand{\dpos}[1]{\cellcolor{okbg}\textcolor{okfg}{#1}}
\newcommand{\dneg}[1]{\cellcolor{nobg}\textcolor{nofg}{#1}}

\newcommand{\dzero}[1]{\cellcolor{gray!10}\textcolor{gray!75}{#1}}

\usepackage{pgfplots}
\pgfplotsset{compat=1.18}

\usepackage{listings}
\lstdefinestyle{promptbox}{
    basicstyle=\ttfamily\scriptsize,
    breaklines=true,
    breakindent=0pt,
    columns=fullflexible,
    keepspaces=true,
    frame=single,
    framesep=2pt,
    rulecolor=\color{gray!55},
    backgroundcolor=\color{gray!6},
    xleftmargin=0pt,
    xrightmargin=0pt,
    aboveskip=4pt,
    belowskip=2pt,
    literate={\{}{{\{}}1 {\}}{{\}}}1,
    escapeinside={(*@}{@*)},
}
\lstnewenvironment{promptquote}{\lstset{style=promptbox}}{}
\usepackage{tcolorbox}

\AtBeginDocument{\pdfstringdefDisableCommands{\def\new#1{#1}}}

\definecolor{cvprblue}{rgb}{0.21,0.49,0.74}
\usepackage[pagebackref,breaklinks,colorlinks,allcolors=cvprblue]{hyperref}

\def\confName{CVPR}
\def\confYear{2027}

\title{
Understanding Multimodal Social Interaction Beyond the Image Plane
}

\author{
Sangho Kim$^{1,2}$ \qquad Georgios Pavlakos$^{2}$ \\
$^{1}$Sungkyunkwan University \quad $^{2}$The University of Texas at Austin \\
}

\begin{document}
\maketitle
\begin{abstract}
Understanding social interactions requires interpreting not only what people say but also how they behave toward one another.
Particularly in multi-party conversations, this becomes highly complex, requiring a holistic understanding of the speaker's spatial and social relationships with surrounding participants.
However, conventional approaches relying on 2D representations often struggle to interpret the speaker's nonverbal cues, such as gaze and gestures, which are directional and thus depend on the participants' spatial arrangement.
We therefore introduce MMSI4D, the first social referent reasoning approach that explicitly encodes the speaker's 4D motion and 3D relationships with each participant.
Leveraging off-the-shelf human mesh recovery, our model represents each of these relationships as a distinct token, allowing it to connect the speaker's cues to each candidate.
On two social referent reasoning tasks of the MMSI benchmark, MMSI4D more than doubles the gain of the prior 2D keypoint-based approach over language-only models and outperforms recent zero-shot video large multimodal models.
Controlled comparisons demonstrate that 3D representations consistently improve accuracy over their 2D projections across architectures, underscoring the value of reasoning explicitly in 3D.
Our analysis further shows that the benefit of 3D grows as participants appearing close in the image lie at increasingly different distances from the camera, highlighting the 3D arrangement as key to resolving the ambiguities inherent in multi-party interactions.
\vspace{-2.5mm}
\end{abstract}    
\section{Introduction}
\label{sec:introduction}

As AI systems increasingly operate alongside people, they need to understand not only what individuals do but also how people interact with one another~\cite{mathur2024advancing,lee2024towards}.
Conversations among several people are a common yet challenging setting for such understanding, since following them requires more than knowing \emph{what} is said.
One must also track \emph{whom} each utterance is directed at and \emph{whom} it mentions, a problem we refer to as \emph{social referent reasoning}.

\begin{figure}[t]
    \centering
    \includegraphics[width=\linewidth]{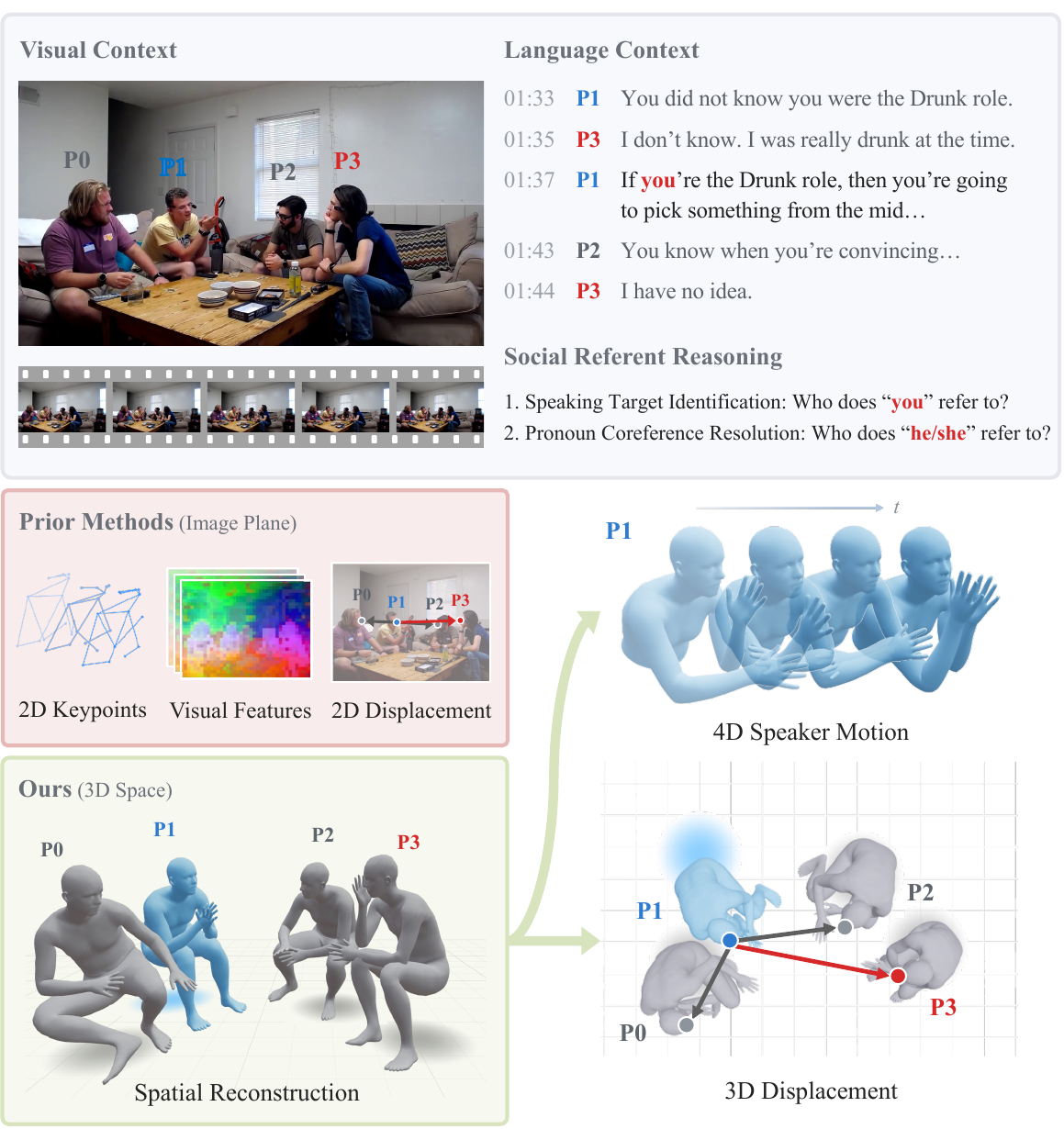}
    \caption{\textbf{Social referent reasoning beyond the image plane.}
    Given the multimodal context of a \new{multi-party} conversation, the task is to identify whom the speaker addresses or mentions.
    \new{Prior methods reason on the image plane}, where the \new{referent} (\textcolor{cTarget}{P3}) and P2 lie in nearly the same direction \new{from} the speaker (\textcolor{cSpeaker}{P1}).
    \new{MMSI4D instead relates the speaker's cues to each participant in the reconstructed 3D space}, where these directions clearly diverge.}    \vspace{+2mm}
    \label{fig:motivation}
    \vspace{-2mm}
\end{figure}

This problem is difficult because conversational expressions are often linguistically underspecified, leaving the listener to infer the intended referent~\cite{clark1986referring,jovanovic2006addressee}.
When a speaker in a group says ``I think you are lying,'' for example, the words alone do not tell who ``you'' is.
The surrounding dialogue can narrow down the candidates, yet it often leaves more than one plausible referent.
People naturally resolve this remaining ambiguity by observing the speaker, as cues like gaze, head orientation, and gestures clearly signal attention~\cite{langton2000you,tanenhaus1995integration,clark2004speaking,hanna2007speakers}.
These cues are inherently directional, which makes the \new{spatial arrangement} of all participants essential for identifying \new{the correct referent}.

Recent work has \new{shown} that modeling the speaker's nonverbal cues helps resolve referents in multi-party conversations.
For instance, MMSI~\cite{lee2024modeling} introduces a multimodal approach that combines linguistic context with these cues, explicitly aligning them with individual participants\new{.}
In MMSI, each participant is represented by the image position of their nose, \new{expressed as a pixel offset from the speaker}.
Subsequent work extends this setting to online reasoning~\cite{li2025towards}, identity attribution from raw audiovisual input~\cite{li2026omni}, and speaker-consistent vision-language alignment~\cite{ouyang2026multi}.
Although these models successfully integrate language and vision, they locate participants in the 2D image plane or encode the scene with holistic visual features\new{, without explicitly modeling the 3D arrangement of participants.}

\new{Such image-plane positions lack each participant's distance from the camera and can thus leave} the target of the speaker's directional cues ambiguous.
In the video frame of Fig.~\ref{fig:motivation}, for instance, the addressee (P3) and another participant (P2) lie in nearly the same direction \new{relative to} the speaker (P1), whereas in 3D their directions clearly differ.
\new{Given only their 2D displacements, the speaker's cues toward this direction could indicate either of them.}
Holistic visual features, \new{on the other hand, might implicitly capture such distances, but they do not directly attribute them to individual participants.}
These challenges underscore the need for an explicit 3D representation of each participant.

In this paper, we introduce MMSI4D, the first approach to social referent reasoning that explicitly encodes the speaker's 4D motion and \new{relates it to} the 3D arrangement of all participants.
Using an off-the-shelf human mesh recovery (HMR) model~\cite{goel2023humans}, we reconstruct and track every participant in 3D from in-the-wild videos of social deduction games.
For each target utterance, MMSI4D encodes the speaker's upper-body motion \new{over a short temporal window}.
For each candidate, it encodes their 3D \new{geometry relative to} the speaker and the speaker's attention toward them.
Unlike prior work that concatenates candidate features into a single token~\cite{lee2024modeling}, MMSI4D assigns each candidate a distinct token, keeping the spatial evidence of different candidates separate.
\new{Finally, these} tokens are fused with the \new{dialogue context} to form a global prediction, to which each token adds a residual correction for its own candidate.
\new{Together, these components help MMSI4D resolve referents that are ambiguous in the image plane but distinguishable in 3D.}

On the YouTube and Ego4D subsets of the MMSI benchmark~\cite{lee2024modeling}, MMSI4D more than doubles the average gain of MMSI over language-only models~\cite{liu2019roberta,devlin-etal-2019-bert,clark2020electra} and outperforms zero-shot video large multimodal models (LMMs) \new{with 7--8B parameters}~\cite{Qwen2.5-VL,bai2025qwen3, zhang2024llava,wang2025internvl3_5}.
Controlled comparisons with 2D projections of the same keypoints show that reasoning in 3D consistently improves accuracy for both MMSI4D and the MMSI architecture.
With the same 3D inputs, \new{our candidate-wise design} outperforms the \new{single-token design} of MMSI.
Further analysis of MMSI4D shows that the advantage of 3D over 2D projections grows as participants \new{appearing} close in the image lie \new{farther apart along the camera axis}.

Our contributions are as follows:
\begin{itemize}
    \item \new{We propose MMSI4D, which is, to our knowledge,} the first approach to social referent reasoning that explicitly encodes the speaker's 4D motion and the \new{3D arrangement} of all participants, both reconstructed from in-the-wild monocular video.
    \item \new{While prior work summarizes candidates in a single token,} we introduce a candidate-wise design that represents and scores each of them separately, reflecting the per-person nature of referent reasoning in multi-party conversations.
    \item We analyze participants' spatial arrangements across two collections of social deduction games and show that \new{the benefit of 3D grows as participants adjacent in the image lie at increasingly different distances from the camera.}
\end{itemize}
\section{Related Work}
\label{sec:related_work}

\subsection{Multimodal Social Interaction Understanding}

Video understanding has long focused on recognizing activities, from individual actions~\cite{carreira2017quo,feichtenhofer2019slowfast,fan2021multiscale,liu2022video} to group activities involving interacting actors~\cite{wu2019learning,li2021groupformer,han2022dual,kim2022detector}.
\new{Beyond activity recognition}, other work studies social behavior~\cite{cao2025socialgesture,lai2023werewolf}, interpersonal relations~\cite{li2024socialgpt}, and \new{question answering}~\cite{zadeh2019social}.

A central question in this direction is who interacts with whom.
Audiovisual models capture speaking and listening relations~\cite{jia2024audio}, \new{while others detect nonverbal interactions} among individuals and groups~\cite{wei2024nonverbal}.
For multi-party conversations, \citet{lee2024modeling} \new{introduce the tasks} of identifying whom each utterance addresses and whom it mentions, and \new{align} each participant's visual cues with their identity in the dialogue.
Subsequent work relaxes the assumptions of this setting by reasoning online without future context~\cite{li2025towards} or directly from raw audiovisual input~\cite{li2026omni}, and strengthens speaker-consistent vision--language alignment in multimodal large language models \new{(MLLMs)}~\cite{ouyang2026multi}.
Related benchmarks further probe addressee recognition~\cite{inoue2025llm}, gaze- and gesture-grounded social reasoning~\cite{kim2026grasp}, and multi-speaker understanding~\cite{chowdhury2026amuse}.

In these \new{works}, participant geometry is given by 2D keypoints~\cite{lee2024modeling}, bounding boxes~\cite{li2026omni,ouyang2026multi}, or visual prompts~\cite{li2025towards}, or remains implicit in visual features~\cite{lai2023werewolf,chowdhury2026amuse}.
None of them, however, represents participants explicitly in 3D, although the speaker's directional cues depend on the participants' \new{spatial} arrangement.
We instead reason over all participants in a shared 3D space and show that this consistently \new{outperforms reasoning in 2D}.
\begin{figure*}[t]
    \centering
    \includegraphics[width=\linewidth]{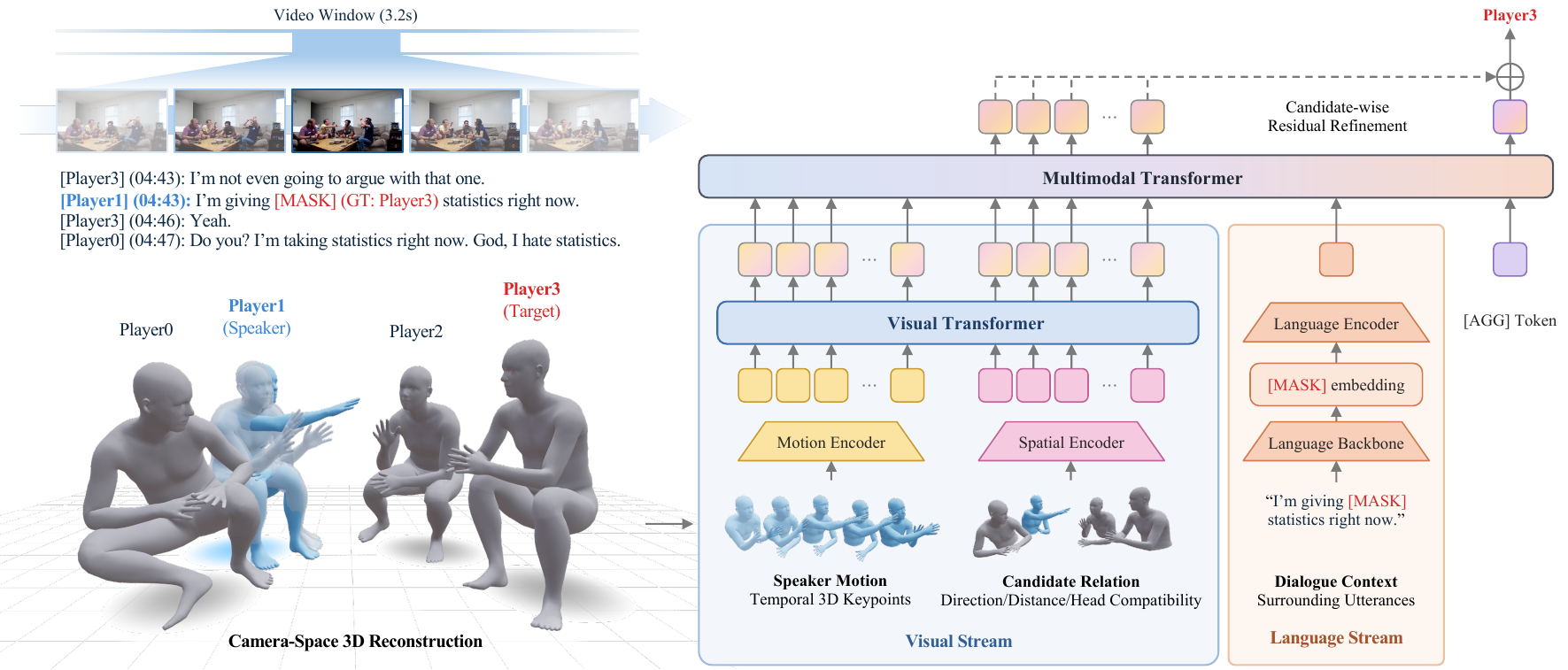}
    \caption{\new{\textbf{Overview of MMSI4D.}}
    Given a video window and its dialogue, where \textcolor{cTarget}{\texttt{[MASK]}} marks the target reference, MMSI4D identifies the participant it denotes.
    \textbf{Left:} All participants are reconstructed in camera space and linked to their identities in the dialogue.
    \textbf{Right:} The \textcolor{cMotion}{speaker motion} tokens encode the speaker's 3D keypoints, one per time step, while the \textcolor{cRelation}{candidate relation} tokens encode each candidate's 3D geometry relative to the speaker, one per candidate.
    Both are then contextualized together by a visual Transformer.
    The \textcolor{cLang}{dialogue context} is encoded into a \textcolor{cTarget}{\texttt{[MASK]}} embedding by a pretrained language model at the \textcolor{cTarget}{\texttt{[MASK]}} position, and a multimodal Transformer fuses both streams.
    The \textcolor{cAgg}{\texttt{[AGG]} token} gives a global prediction, which candidate-wise residual refinement adjusts for each candidate.}
    \label{fig:architecture}
    \vspace{-1mm}
\end{figure*}

\subsection{3D Human Geometry for Interaction Modeling}
 
Human pose estimation has progressed from localizing image-space keypoints~\cite{cao2017realtime,fang2022alphapose,xu2022vitpose} to recovering articulated 3D bodies with parametric models such as SMPL~\cite{loper2023smpl} and SMPL-X~\cite{pavlakos2019expressive}.
For a single person, monocular methods reconstruct such bodies from images~\cite{kanazawa2018end,kolotouros2019learning,kocabas2021pare} and videos~\cite{kocabas2020vibe,choi2021beyond}, and some \new{also} place them in world coordinates~\cite{shin2024wham}.
For multiple people, \new{other} methods reconstruct all individuals in a shared camera space from a single image~\cite{jiang2020coherent,baradel2024multi} and track them over time in video~\cite{rajasegaran2022tracking,goel2023humans}.
\new{Recent methods further reconstruct the surrounding scene together with humans} in a shared 3D coordinate frame~\cite{chen2025human3r,muller2025reconstructing}.
 
These 3D representations have enabled modeling of how people interact with scenes~\cite{ashutosh2025fiction,jiang2024scaling,li2024genzi,wang2021synthesizing}, objects~\cite{dwivedi_interactvlm_2025,huang2026recovering}, and other people~\cite{huang2024closely,maluleke2024synergy,muller2024generative,zheng2025efficient}.
For social behavior, 3D humans have been used to capture close physical contact~\cite{khirodkar2024harmony4d} and large-scale multimodal behavior~\cite{mclean2025embody}, to generate multi-person motion from language~\cite{huang2026unified,yu2026socialgen}, and to predict nonverbal signals such as speaking status and social formation in triadic interactions~\cite{joo2019towards}.
\new{However, most of these works focus on reconstructing or generating 3D humans, while those using them as input for social inference rely on multi-camera capture in a controlled setting~\cite{joo2019towards}.}
\new{We instead obtain this input from in-the-wild monocular video with an off-the-shelf model~\cite{goel2023humans} and use it to infer whom an utterance addresses or mentions.}
 
\subsection{Directional Cues for Referential Reasoning}
 
People infer what others attend to from directional cues such as gaze, head orientation, and pointing~\cite{emery2000eyes,langton2000you}.
Early work used head orientation to estimate attention in meetings~\cite{ba2008recognizing,stiefelhagen2002modeling} and, together with speech, to infer addressees in conversations among humans, agents, and robots~\cite{otsuka2005probabilistic,huang2011making,katzenmaier2004identifying}.
In computer vision, gaze has since been studied for target detection~\cite{chong2020detecting,gupta2024mtgs,ryan2025gaze,tonini2023object,tu2022end}, joint attention~\cite{fan2018inferring,nakatani2023interaction}, and gaze communication~\cite{fan2019understanding}, and recent methods further incorporate 3D scene geometry~\cite{bao2022escnet,fang2021dual,kawana2025ga3ce,yun2026gaze}.
 
Gesture research has likewise progressed from recognizing bodily behaviors in social contexts~\cite{balazia2022bodily,cao2025socialgesture} to estimating the targets of deictic gestures such as pointing~\cite{cao2026toward}.
Recent work combines pointing with language to ground referred objects in 3D or egocentric scenes~\cite{mane2025ges3vig,li2026beyond,li2026pointvg} and to answer pointing-based questions with MLLMs~\cite{choi2026you,li2026mllms}.
Among them, methods that explicitly \new{model} hand keypoints or the pointing direction~\cite{choi2026you,li2026pointvg} show that such geometric cues help identify the referent of a gesture.

\new{Most of these methods treat image regions, scene locations, or objects as targets, while early addressee estimation was mostly studied in constrained settings.}
We instead consider in-the-wild group conversations, where the targets of the speaker's directional cues are the other participants.
\new{With these participants reconstructed in 3D, we directly compare the speaker's head orientation with the direction toward each of them and relate the speaker's gestures to their positions, obtaining per-participant evidence of attention.}

\section{Method}
\label{sec:method}

\subsection{Problem Formulation and Overview}
\label{sec:method_overview}

We address two social referent reasoning tasks introduced by \citet{lee2024modeling} and adopted in subsequent work~\cite{li2025towards,li2026omni}.
Speaking Target Identification (STI) identifies the addressee of a second-person expression such as ``you'' or ``your,'' and Pronoun Coreference Resolution (PCR) identifies the participant referred to by a third-person pronoun such as ``he'' or ``her.''
\new{Both tasks are formulated as classifying a \texttt{[MASK]} token in the conversation into one of the $N$ participants.}
Given the dialogue context $\mathcal{U}_k=\{u_i\}_{i=k-n}^{k+n}$ centered at the target utterance $u_k$ and the video $\mathcal{V}=\{v_t\}_{t=1}^{T}$ temporally aligned with $u_k$, we predict
\begin{equation}
    \hat{y}=\arg\max_{i\in\{1,\ldots,N\}} p\left(y=i \mid \mathcal{U}_k,\mathcal{V}\right).
    \label{eq:problem_formulation}
\end{equation}

\new{MMSI4D identifies the referent by relating the speaker's nonverbal behavior to the location of each candidate (Fig.~\ref{fig:architecture}).}
\new{To this end, we reconstruct all participants in camera space (Sec.~\ref{sec:3d_reconstruction}) and use these reconstructions to encode the speaker's 4D motion (Sec.~\ref{sec:speaker_motion}) and the 3D geometry of each candidate relative to the speaker (Sec.~\ref{sec:candidate_geometry}).}
\new{We then embed each candidate's geometry as a separate token and contextualize these tokens with the speaker's motion (Sec.~\ref{sec:candidate_encoding}).}
\new{The resulting tokens are fused with the dialogue context into a global prediction, which is finally refined through residual corrections predicted from each candidate's output token (Sec.~\ref{sec:multimodal_fusion}).}
\new{We adopt the language backbone, Transformer configuration, and language-visual alignment of MMSI~\cite{lee2024modeling} for a controlled comparison.}

\subsection{3D Human Reconstruction}
\label{sec:3d_reconstruction}

We first reconstruct and track all participants in $\mathcal{V}$ with 4DHumans~\cite{goel2023humans}, which combines HMR2.0 mesh recovery with 3D-aware tracking~\cite{rajasegaran2022tracking}.
For participant $i$ at frame $t$, HMR2.0 predicts each joint $j$ in body-centered coordinates, $\tilde{\mathbf{p}}^{j}_{i,t}\in\mathbb{R}^{3}$, together with the body translation $\boldsymbol{\tau}_{i,t}\in\mathbb{R}^{3}$ in camera coordinates, from which we obtain camera-space joints $\mathbf{p}^{j}_{i,t}=\tilde{\mathbf{p}}^{j}_{i,t}+\boldsymbol{\tau}_{i,t}$.
\new{These joints are then mapped to the 17 COCO body keypoints.}
\new{Finally, each} track is linked to its player identity in the dialogue to locate the speaker $s$ and each candidate, extending the language-visual alignment~\cite{lee2024modeling} \new{to 3D tracks}.

\subsection{\new{4D Speaker Motion}}
\label{sec:speaker_motion}

We describe the speaker's nonverbal behavior with a set $\mathcal{J}$ of nine head and upper-body joints, \new{namely the nose, eyes, shoulders, elbows, and wrists, which together capture head orientation and arm gestures}.
At each frame, we express every joint relative to the speaker's nose, $\mathbf{q}^{j}_{t}=\mathbf{p}^{j}_{s,t}-\mathbf{p}^{\mathrm{nose}}_{s,t}$, which factors out where the speaker is located while preserving the 3D configuration of the head and arms.
We then embed each joint with a point encoder $E_{\mathrm{point}}$ shared across joints, and map the concatenated joint embeddings to a per-frame feature $\mathbf{f}_{t}=E_{\mathrm{motion}}\big([E_{\mathrm{point}}(\mathbf{q}^{j}_{t})]_{j\in\mathcal{J}}\big)$ with a motion encoder $E_{\mathrm{motion}}$.
We subsample these features to $T'$ time steps and add sinusoidal positional encoding~\cite{vaswani2017attention}, yielding motion tokens $\mathbf{M}=[\mathbf{m}_1,\ldots,\mathbf{m}_{T'}]$ that preserve the temporal order of the speaker's behavior.

\subsection{\new{Speaker--Candidate 3D Geometry}}
\label{sec:candidate_geometry}

We next describe each candidate $i$ by their position relative to the speaker, so that the speaker's motion can be related to specific participants.
Taking the nose as each participant's \new{position}, we compute the \new{3D} displacement $\Delta\mathbf{p}_{i,t}=\mathbf{p}^{\mathrm{nose}}_{i,t}-\mathbf{p}^{\mathrm{nose}}_{s,t}$ over the set $\mathcal{T}_i$ of frames in which both the candidate and the speaker are detected, and take its temporal median to suppress per-frame \new{reconstruction} noise,
\begin{equation}
    \mathbf{r}_i=\Pi_{XZ}\Big(\operatorname*{median}_{t\in\mathcal{T}_i}\Delta\mathbf{p}_{i,t}\Big)\in\mathbb{R}^{2},
    \label{eq:relative_position}
\end{equation}
where $\Pi_{XZ}(\cdot)$ denotes the projection onto the camera's $XZ$ plane.
From $\mathbf{r}_i$, we further derive the unit direction $\mathbf{u}_i=\mathbf{r}_i/\max(\|\mathbf{r}_i\|_2,\epsilon)$ and the distance $d_i=\|\mathbf{r}_i\|_2$.

\paragraph{Head--candidate compatibility.}
\new{The relative position $\mathbf{r}_i$, however, only represents where each candidate is, not how much the speaker attends to them.}
We therefore add a compatibility score $c_i$ that measures how directly the speaker faces each candidate, \new{as head orientation signals the direction of attention}~\cite{langton2000you}.
We \new{approximate} the speaker's head orientation by the direction from the eye midpoint to the nose, $\mathbf{g}_t=\Pi_{XZ}(\mathbf{p}^{\mathrm{nose}}_{s,t}-\bar{\mathbf{p}}^{\mathrm{eye}}_{s,t})$ with $\bar{\mathbf{p}}^{\mathrm{eye}}_{s,t}=(\mathbf{p}^{\mathrm{l\text{-}eye}}_{s,t}+\mathbf{p}^{\mathrm{r\text{-}eye}}_{s,t})/2$.
We then compare it with the direction toward the candidate by cosine similarity and take the maximum over \new{$\mathcal{T}_i$}, so that even a brief turn toward the candidate is retained,
\begin{equation}
    c_i=\max_{t\in\mathcal{T}_i}
    \frac{\mathbf{g}_t^{\top}\Pi_{XZ}(\Delta\mathbf{p}_{i,t})}
         {\|\mathbf{g}_t\|_2\,\|\Pi_{XZ}(\Delta\mathbf{p}_{i,t})\|_2}.
    \label{eq:head_compat}
\end{equation}
Finally, a binary indicator $s_i=\mathbf{1}[i=s]$ \new{marks whether candidate $i$ is the speaker}, and these \new{quantities} are concatenated into the relation feature of candidate $i$,
\begin{equation}
    \mathbf{z}_i=[\mathbf{r}_i;\mathbf{u}_i;d_i;s_i;c_i]\in\mathbb{R}^{7}.
    \label{eq:candidate_relation}
\end{equation}

\begin{table*}[t]
\centering
\caption{\textbf{Main results} on Speaking Target Identification (STI) and Pronoun Coreference Resolution (PCR) on the YouTube and Ego4D subsets.
We report accuracy (\%), with Avg.\ denoting the unweighted mean of the two subsets, computed before rounding.
Video LMMs are evaluated zero-shot, and results of the task-specific baselines are taken from~\cite{lee2024modeling}.
\textbf{Bold} and \underline{underline} indicate the best and second-best results within each language-backbone group.}
\label{tab:main_results}

\scriptsize
\setlength{\tabcolsep}{3.3pt}

\resizebox{\textwidth}{!}{
\begin{tabular}{l cccc ccc ccc}
\toprule
& \multicolumn{4}{c}{\textbf{Representation / Modeling}}
& \multicolumn{3}{c}{\textbf{Speaking Target Identification}}
& \multicolumn{3}{c}{\textbf{Pronoun Coreference Resolution}} \\
\cmidrule(lr){2-5} \cmidrule(lr){6-8} \cmidrule(lr){9-11}
\textbf{Method} &
\shortstack{\textbf{Aligned?}} & \textbf{Explicit?} & \textbf{3D?} & \shortstack{\textbf{Candidate-wise?}} &
\textbf{YouTube} & \textbf{Ego4D} & \textbf{Avg.} &
\textbf{YouTube} & \textbf{Ego4D} & \textbf{Avg.} \\
\midrule
\multicolumn{11}{l}{\textit{Video LMMs (zero-shot)}} \\[1pt]
LLaVA-Video-7B~\cite{zhang2024llava}   & -- & \textcolor{red}{\xmark} & \textcolor{red}{\xmark} & -- & 54.8 & 52.3 & 53.5 & 54.7 & 41.1 & 47.9 \\
Qwen2.5-VL-7B~\cite{Qwen2.5-VL}        & -- & \textcolor{red}{\xmark} & \textcolor{red}{\xmark} & -- & 61.5 & 58.0 & 59.7 & 69.8 & 52.7 & 61.2 \\
InternVL3.5-8B~\cite{wang2025internvl3_5} & -- & \textcolor{red}{\xmark} & \textcolor{red}{\xmark} & -- & 68.5 & 60.8 & 64.7 & 56.2 & 47.3 & 51.8 \\
Qwen3-VL-8B~\cite{bai2025qwen3} & -- & \textcolor{red}{\xmark} & \textcolor{red}{\xmark} & -- & 69.0 & 65.3 & 67.2 & 61.0 & 47.3 & 54.2 \\
\midrule
\multicolumn{11}{l}{\textit{Task-specific multimodal models}} \\[1pt]
BERT~\cite{devlin-etal-2019-bert} & \textcolor{red}{\xmark} & \textcolor{red}{\xmark} & \textcolor{red}{\xmark} & \textcolor{red}{\xmark} & 65.8 & 56.8 & 61.3 & 60.3 & 47.3 & 53.8 \\
BERT + DINOv2~\cite{oquab2023dinov2} & \textcolor{red}{\xmark} & \textcolor{red}{\xmark} & \textcolor{red}{\xmark} & \textcolor{red}{\xmark} & 66.4 & 58.0 & 62.2 & 58.2 & 46.4 & 52.3 \\
BERT + MViT~\cite{fan2021multiscale,lai2023werewolf} & \textcolor{red}{\xmark} & \textcolor{red}{\xmark} & \textcolor{red}{\xmark} & \textcolor{red}{\xmark} & 66.9 & 57.4 & 62.2 & 59.8 & 46.4 & 53.1 \\
BERT-based MMSI~\cite{lee2024modeling} & \textcolor{green!70!black}{\checkmark} & \textcolor{green!70!black}{\checkmark} & \textcolor{red}{\xmark} & \textcolor{red}{\xmark}
& \underline{72.7} & \underline{61.9} & \underline{67.3} & \underline{65.9} & \underline{49.1} & \underline{57.5} \\
\textbf{BERT-based MMSI4D (Ours)} & \textcolor{green!70!black}{\checkmark} & \textcolor{green!70!black}{\checkmark} & \textcolor{green!70!black}{\checkmark} & \textcolor{green!70!black}{\checkmark}
& \textbf{74.2} & \textbf{75.6} & \textbf{74.9} & \textbf{69.0} & \textbf{58.9} & \textbf{64.0} \\
\cmidrule(lr){1-11}
RoBERTa~\cite{liu2019roberta} & \textcolor{red}{\xmark} & \textcolor{red}{\xmark} & \textcolor{red}{\xmark} & \textcolor{red}{\xmark} & 72.4 & 63.6 & 68.0 & 69.0 & 48.2 & 58.6 \\
RoBERTa + DINOv2~\cite{oquab2023dinov2} & \textcolor{red}{\xmark} & \textcolor{red}{\xmark} & \textcolor{red}{\xmark} & \textcolor{red}{\xmark} & 72.7 & 62.5 & 67.6 & 68.6 & 46.4 & 57.5 \\
RoBERTa + MViT~\cite{fan2021multiscale,lai2023werewolf} & \textcolor{red}{\xmark} & \textcolor{red}{\xmark} & \textcolor{red}{\xmark} & \textcolor{red}{\xmark} & 73.1 & 64.2 & 68.7 & 69.5 & 49.1 & 59.3 \\
RoBERTa-based MMSI~\cite{lee2024modeling} & \textcolor{green!70!black}{\checkmark} & \textcolor{green!70!black}{\checkmark} & \textcolor{red}{\xmark} & \textcolor{red}{\xmark}
& \underline{74.5} & \underline{66.5} & \underline{70.5} & \underline{73.0} & \underline{52.7} & \underline{62.9} \\
\textbf{RoBERTa-based MMSI4D (Ours)} & \textcolor{green!70!black}{\checkmark} & \textcolor{green!70!black}{\checkmark} & \textcolor{green!70!black}{\checkmark} & \textcolor{green!70!black}{\checkmark}
& \textbf{74.7} & \textbf{76.7} & \textbf{75.7} & \textbf{75.0} & \textbf{60.7} & \textbf{67.8} \\
\cmidrule(lr){1-11}
ELECTRA~\cite{clark2020electra} & \textcolor{red}{\xmark} & \textcolor{red}{\xmark} & \textcolor{red}{\xmark} & \textcolor{red}{\xmark} & 65.8 & 60.8 & 63.3 & 62.5 & 44.6 & 53.6 \\
ELECTRA + DINOv2~\cite{oquab2023dinov2} & \textcolor{red}{\xmark} & \textcolor{red}{\xmark} & \textcolor{red}{\xmark} & \textcolor{red}{\xmark} & 65.3 & 60.2 & 62.8 & 61.1 & 42.9 & 52.0 \\
ELECTRA + MViT~\cite{fan2021multiscale,lai2023werewolf} & \textcolor{red}{\xmark} & \textcolor{red}{\xmark} & \textcolor{red}{\xmark} & \textcolor{red}{\xmark} & 64.6 & 60.8 & 62.7 & 62.1 & 43.8 & 53.0 \\
ELECTRA-based MMSI~\cite{lee2024modeling} & \textcolor{green!70!black}{\checkmark} & \textcolor{green!70!black}{\checkmark} & \textcolor{red}{\xmark} & \textcolor{red}{\xmark}
& \underline{69.6} & \underline{64.8} & \underline{67.2} & \underline{67.6} & \underline{46.4} & \underline{57.0} \\
\textbf{ELECTRA-based MMSI4D (Ours)} & \textcolor{green!70!black}{\checkmark} & \textcolor{green!70!black}{\checkmark} & \textcolor{green!70!black}{\checkmark} & \textcolor{green!70!black}{\checkmark}
& \textbf{72.1} & \textbf{71.6} & \textbf{71.8} & \textbf{69.4} & \textbf{54.5} & \textbf{61.9} \\
\bottomrule
\end{tabular}
}
\vspace{-2mm}
\end{table*}

\subsection{Candidate-wise Tokenization}
\label{sec:candidate_encoding}

To keep each piece of spatial evidence tied to the participant it describes, we encode each relation feature $\mathbf{z}_i$ as a separate relation token.
We embed $\mathbf{z}_i$ with a two-layer MLP $E_{\mathrm{rel}}$ and add a learned \new{embedding $\mathbf{e}_i$ of its slot index},

\begin{equation}
\mathbf{h}^{\mathrm{rel}}_i=E_{\mathrm{rel}}(\mathbf{z}_i)+\mathbf{e}_i.
    \label{eq:relation_token}
\end{equation}
A visual Transformer $E_{\mathrm{vis}}$ then jointly contextualizes the relation tokens $\mathbf{H}=[\mathbf{h}^{\mathrm{rel}}_1,\ldots,\mathbf{h}^{\mathrm{rel}}_N]$ and the motion tokens $\mathbf{M}$ into $\hat{\mathcal{V}}=E_{\mathrm{vis}}([\mathbf{H};\mathbf{M}])$, so that each candidate's relation is interpreted in light of the speaker's motion.

\subsection{Multimodal Fusion and Residual Refinement}
\label{sec:multimodal_fusion}

We encode the conversational context~$\mathcal{U}_k$ with a pretrained language \new{backbone}, namely BERT~\cite{devlin-etal-2019-bert}, RoBERTa~\cite{liu2019roberta}, or ELECTRA~\cite{clark2020electra}, and project its hidden state at the \texttt{[MASK]} position to a conversation feature $\mathbf{h}^{\mathrm{lang}}$ \new{with a language encoder $E_{\mathrm{lang}}$}.
A multimodal Transformer $E_{\mathrm{multi}}$ then jointly processes $[\mathbf{h}^{\mathrm{agg}};\mathbf{h}^{\mathrm{lang}};\hat{\mathcal{V}}]$, where $\mathbf{h}^{\mathrm{agg}}$ is a learnable aggregation token.
Its output \new{at the aggregation token} summarizes the global multimodal context and is mapped to global logits $o^{g}_i$ by a linear classifier.
\new{To complement this pooled summary with candidate-specific evidence, we further predict a correction to each candidate's logit from the output of its relation token, $\mathbf{h}^{\mathrm{cand}}_i$, using a shared linear scorer $f_{\mathrm{res}}$},
\begin{equation}
o_i=o^{g}_i+\underbrace{f_{\mathrm{res}}\big(\mathbf{h}^{\mathrm{cand}}_i\big)}_{\delta_i}.
\label{eq:residual}
\end{equation}
We initialize $f_{\mathrm{res}}$ to zero so that training begins from the global prediction, \new{which the residual branch then gradually learns to refine}.

\subsection{Training}
\label{sec:objective}

\new{We assign participants to $C=6$ slots, the maximum number of participants per conversation in the benchmark, and pad and mask unused slots.}
\new{We train the model} with the cross-entropy loss on the final logits $\{o_i\}_{i=1}^{C}$.
\new{During training,} we randomly shuffle the mapping from participants to slots and remap the \new{player identifiers in the dialogue, the} labels, and \new{the} observations accordingly, which prevents the model from relying on fixed slot indices.
We \new{also} randomly mirror the reconstructed geometry by negating its $X$ coordinates, which encourages robustness to left--right variation while leaving all distances and head compatibilities unchanged.
\section{Experiments}
\label{sec:experiments}

\begin{table}[t]
\centering
\caption{\textbf{2D vs.\ 3D inputs from the same pose estimator.}
Both use the same HMR2.0~\cite{goel2023humans} keypoints, either projected onto the image plane (2D) or kept in 3D.
We report accuracy (\%) averaged over both subsets, with $\Delta$ denoting the gain of 3D over 2D.
Per-subset results of MMSI4D are provided in the supplementary material.
}
\label{tab:representation_study}
\scriptsize
\setlength{\tabcolsep}{4pt}
\resizebox{\linewidth}{!}{
\begin{tabular}{l ccc ccc}
\toprule
& \multicolumn{3}{c}{STI} & \multicolumn{3}{c}{PCR} \\
\cmidrule(lr){2-4}\cmidrule(lr){5-7}
Backbone & 2D & 3D & $\Delta$ & 2D & 3D & $\Delta$ \\
\midrule
\multicolumn{7}{l}{\textit{MMSI architecture~\cite{lee2024modeling}}} \\
BERT~\cite{devlin-etal-2019-bert}    & 68.8 & 70.5 & \dpos{+1.7} & 57.8 & 62.9 & \dpos{+5.1} \\
RoBERTa~\cite{liu2019roberta}        & 72.7 & 73.8 & \dpos{+1.1} & 65.1 & 66.2 & \dpos{+1.1} \\
ELECTRA~\cite{clark2020electra}      & 69.0 & 72.1 & \dpos{+3.1} & 59.3 & 59.9 & \dpos{+0.6} \\
\midrule
\multicolumn{7}{l}{\textit{MMSI4D (ours)}} \\
BERT~\cite{devlin-etal-2019-bert}    & \new{70.5} & 74.9 & \dpos{\new{+4.4}} & \new{60.7} & 64.0 & \dpos{\new{+3.3}} \\
RoBERTa~\cite{liu2019roberta}    & 72.9 & 75.7 & \dpos{+2.8} & 64.7 & 67.8 & \dpos{+3.1} \\
ELECTRA~\cite{clark2020electra}      & 68.1 & 71.8 & \dpos{+3.7} & 60.0 & 61.9 & \dpos{+1.9} \\
\bottomrule
\end{tabular}}
\vspace{-2mm}
\end{table}

\begin{table*}[ht!]
\centering
\caption{\textbf{Component ablation of MMSI4D} with the BERT~\cite{devlin-etal-2019-bert} backbone.
\replmark{}:~the input is computed from the projected 2D keypoints.
\nomark{}:~the component is removed.
\textbf{Bold} and \underline{underline} indicate the best and second-best results.}
\label{tab:main_ablation}

\scriptsize
\setlength{\tabcolsep}{4.5pt}
\renewcommand{\arraystretch}{1.15}

\resizebox{\textwidth}{!}{
\begin{tabular}{
    l
    cccc|
    ccc
    ccc
}
\toprule

\multirow[c]{2}{*}{\textbf{Variant}} &
\multicolumn{4}{c|}{\textbf{Model Components}} &
\multicolumn{3}{c}{\textbf{Speaking Target Identification}} &
\multicolumn{3}{c}{\textbf{Pronoun Coreference Resolution}} \\

\cmidrule(lr){2-5}
\cmidrule(lr){6-8}
\cmidrule(lr){9-11}

&
\shortstack{\textbf{Speaker}\\\textbf{Motion}} &
\shortstack{\textbf{Relative}\\\textbf{Position}} &
\shortstack{\textbf{Head--Candidate}\\\textbf{Compatibility}} &
\multicolumn{1}{c|}{\shortstack{\textbf{Residual}\\\textbf{Refinement}}} &
\textbf{YouTube} & \textbf{Ego4D} & \textbf{Avg.} &
\textbf{YouTube} & \textbf{Ego4D} & \textbf{Avg.} \\

\midrule

\textbf{MMSI4D (Full)}
& \yes & \yes & \yes & \yes
& \textbf{74.2} & \textbf{75.6} & \textbf{74.9}
& \underline{69.0} & \textbf{58.9} & \textbf{64.0} \\

\midrule

\hspace{1em}w/ 2D Speaker Motion
& \repl & \yes & \yes & \yes
& 72.2 & 68.8 & 70.5
& \textbf{69.6} & 52.7 & 61.1 \\

\hspace{1em}w/ 2D Relative Position
& \yes & \repl & \yes & \yes
& 71.9 & \underline{70.5} & \underline{71.2}
& 66.3 & 50.0 & 58.2 \\

\hspace{1em}w/o Head--Candidate Compatibility
& \yes & \yes & \no & \yes
& 71.1 & 69.3 & 70.2
& 67.9 & \underline{54.5} & \underline{61.2} \\

\hspace{1em}w/o Residual Refinement
& \yes & \yes & \yes & \no
& \underline{73.4} & 68.8 & 71.1
& 66.0 & 53.6 & 59.8 \\

\bottomrule
\end{tabular}
}
\vspace{-2mm}
\end{table*}

\subsection{Experimental Setup}
\label{sec:setup}

\paragraph{Datasets and metrics.}
We evaluate on the YouTube and Ego4D~\cite{grauman2022ego4d} subsets of \textit{Werewolf Among Us}~\cite{lai2023werewolf}, which record multi-party social deduction games, using the STI and PCR annotations of MMSI~\cite{lee2024modeling}.
\new{The YouTube subset includes 151 videos (14.8\,h) with 3--6 participants and provides 3,255 STI and 2,679 PCR instances, while the Ego4D subset includes 83 videos (6.0\,h) with 4--5 participants and provides 832 STI and 503 PCR instances.}
Following prior work~\cite{lee2024modeling,li2025towards,li2026omni}, we split each subset into training and test sets at the video level and report accuracy (\%) on each subset.
We also report the unweighted average over the two subsets.

\paragraph{Baselines.}
We compare with three groups of task-specific models built on the same language encoders~\cite{devlin-etal-2019-bert,liu2019roberta,clark2020electra}, namely language-only models, models augmented with holistic visual features from \new{DINOv2~\cite{oquab2023dinov2} or MViT~\cite{fan2021multiscale} as in~\cite{lai2023werewolf}}, and MMSI~\cite{lee2024modeling}, which uses participant-aligned 2D keypoints.
\new{We report their published results~\cite{lee2024modeling}, as we follow the same data splits and evaluation protocol.}
\new{Subsequent methods address different settings, such as online reasoning without future context~\cite{li2025towards} and identity attribution from raw audiovisual input~\cite{li2026omni}, and are thus not directly comparable.}
\new{We further evaluate the 7--8B variants of four recent open-weight video LMMs~\cite{zhang2024llava,Qwen2.5-VL,wang2025internvl3_5,bai2025qwen3} in a zero-shot setting, providing the same dialogue context and participant identities as our model.}

\paragraph{Implementation details.}
For each target utterance, we use the five preceding and five following utterances as context and a 16-frame visual window (3.2\,s) around the utterance, subsampled into eight motion tokens.
All modules other than the language encoder use a hidden size of 512, with three Transformer layers in $E_{\mathrm{vis}}$ and two in $E_{\mathrm{multi}}$.
We train with Adam~\cite{kingma2014adam} for 200 epochs using a batch size of 16 and learning rates of $5\times10^{-6}$ for the language encoder and $5\times10^{-5}$ for all other modules, as in MMSI~\cite{lee2024modeling}.
During the first 10 epochs, the visual branch is warmed up without the language input.
Further details, including the prompts for video LMMs, are provided in the supplementary material.
 
\subsection{Comparison with Prior Methods}
\label{sec:main_results}

As shown in Table~\ref{tab:main_results}, MMSI4D achieves the best average accuracy on both tasks with all three language backbones, \new{outperforming all compared methods, including MMSI~\cite{lee2024modeling} by a clear margin}.
\new{The gains are larger on the Ego4D subset, where image-plane neighbors differ more in their distances from the camera, as we analyze in Sec.~\ref{sec:when_3d_helps} and the supplementary material.}

Among the compared models, approaches that \new{use} participant-aligned keypoints, namely MMSI and MMSI4D, gain considerably more over the language-only models than those augmented with holistic visual features from DINOv2~\cite{oquab2023dinov2} or MViT~\cite{fan2021multiscale}.
This suggests that \new{encoding directional nonverbal cues explicitly for each participant is more effective than leaving them implicit} in global visual features.
\new{Between these two explicit methods, the average gain of MMSI4D is more than twice that of MMSI (9.6 vs.\ 4.0 points).
We disentangle the factors contributing to this gain in Secs.~\ref{sec:effect_3d}--\ref{sec:ablations}.
}

Notably, \new{the best-performing video LMM baselines achieve accuracy comparable to that of MMSI}, highlighting the general multimodal \new{capabilities} acquired through large-scale pretraining.
Nevertheless, MMSI4D outperforms all of them in average accuracy with every backbone, indicating that \new{explicit 3D geometry aligned with each participant remains beneficial} even against such pretrained models.

\subsection{Effect of 3D Spatial Inputs}
\label{sec:effect_3d}

\new{MMSI4D differs from MMSI~\cite{lee2024modeling} not only in architecture but also in the pose estimator, using 3D keypoints from HMR2.0~\cite{goel2023humans} rather than 2D keypoints from AlphaPose~\cite{fang2022alphapose}.}
\new{To evaluate whether its improvement comes from reasoning in 3D, we fix the estimator to HMR2.0 and compare MMSI4D with its 2D variant trained on the projected keypoints.}
We repeat this comparison with the MMSI architecture to examine whether the benefit of 3D generalizes beyond our design.
\new{We minimally modify each model to accept the other input type, as MMSI and MMSI4D are designed for 2D and 3D keypoints, respectively.}
For MMSI, we use its official training code and only extend its keypoint input with the $Z$ coordinate, and for MMSI4D, we compute its motion and relation features from the projected 2D keypoints.

As shown in Table~\ref{tab:representation_study}, 3D inputs consistently yield higher average accuracy than their \new{2D projections} across both architectures, all three backbones, and both tasks.
\new{With both inputs drawn from the same HMR2.0 reconstruction}, these results show that reasoning in 3D improves accuracy \new{independently of} the change in pose estimator and that this gain is not specific to our architecture.

\begin{table}[t]
\centering
\caption{\textbf{Effect of candidate-wise tokenization} with the BERT~\cite{devlin-etal-2019-bert} backbone.
All variants use the same 3D inputs.}
\label{tab:candidate_modeling}
\scriptsize
\setlength{\tabcolsep}{4pt}
\resizebox{\linewidth}{!}{
\begin{tabular}{l ccc ccc}
\toprule
& \multicolumn{3}{c}{STI} & \multicolumn{3}{c}{PCR} \\
\cmidrule(lr){2-4}\cmidrule(lr){5-7}
Variant & YouTube & Ego4D & Avg. & YouTube & Ego4D & Avg. \\
\midrule
Global token                 & 72.5 & 67.0 & 69.8 & 65.2 & \underline{53.6} & 59.4 \\
Candidate-wise token         & \underline{73.4} & \underline{68.8} & \underline{71.1} & \underline{66.0} & \underline{53.6} & \underline{59.8} \\
\hspace{1em}+ Residual ref. (Full) & \textbf{74.2} & \textbf{75.6} & \textbf{74.9} & \textbf{69.0} & \textbf{58.9} & \textbf{64.0} \\
\bottomrule
\end{tabular}}
\end{table}
\begin{table}[t]
\centering
\caption{\textbf{Effect of gaze- and gesture-related cues} with the BERT~\cite{devlin-etal-2019-bert} backbone.
Accuracy (\%) on each subset and their average.
\new{The last row denotes the language-only baseline from Table~\ref{tab:main_results}, shown for reference.}}
\label{tab:headbody}
\scriptsize
\setlength{\tabcolsep}{4pt}
\resizebox{\linewidth}{!}{
\begin{tabular}{l ccc ccc}
\toprule
& \multicolumn{3}{c}{STI} & \multicolumn{3}{c}{PCR} \\
\cmidrule(lr){2-4}\cmidrule(lr){5-7}
Variant & YouTube & Ego4D & Avg. & YouTube & Ego4D & Avg. \\
\midrule
MMSI4D (Full) & \textbf{74.2} & \textbf{75.6} & \textbf{74.9} & \textbf{69.0} & \textbf{58.9} & \textbf{64.0} \\
\midrule
\hspace{1em}w/o Gesture Features & \underline{71.9} & \underline{71.6} & \underline{71.8} & \underline{66.7} & \underline{56.3} & \underline{61.5} \\
\hspace{1em}w/o Gaze Features    & 71.2 & 70.5 & 70.8 & 64.6 & 54.5 & 59.6 \\
\midrule
\hspace{1em}w/o Visual Features  & 65.8 & 56.8 & 61.3 & 60.3 & 47.3 & 53.8 \\
\bottomrule
\end{tabular}}
\vspace{-2mm}
\end{table}

\subsection{Effect of Candidate-wise Tokenization}
\label{sec:candidate_wise}

\new{We next examine the effect of encoding each relation feature $\mathbf{z}_i$ (Eq.~\eqref{eq:candidate_relation}) as a separate token, which keeps each candidate's relation to the speaker distinct.}
For comparison, we concatenate the relation features of all candidates into a single global token, $\mathbf{h}^{\mathrm{glob}}=E_{\mathrm{glob}}([\mathbf{z}_1;\allowbreak\ldots;\allowbreak\mathbf{z}_N])$, following the formulation of MMSI~\cite{lee2024modeling}.
\new{We compare this variant with candidate-wise tokens without residual refinement, as the global token provides no per-candidate outputs for the shared scorer of Eq.~\eqref{eq:residual}.}
Note that all variants receive the same 3D inputs, \new{so the difference between these two variants arises only from how these features are tokenized.}
As shown in Table~\ref{tab:candidate_modeling}, candidate-wise tokens \new{outperform the global token in average accuracy on both tasks.}
For reference, we also report the accuracy of the full model, which adds residual refinement and further improves performance.
\new{Together, these results indicate that candidate-wise tokens benefit referent reasoning both by keeping the evidence for each candidate separate and by enabling residual refinement.}

\subsection{Ablation Studies}
\label{sec:ablations}

\paragraph{\new{Main components.}}
Table~\ref{tab:main_ablation} ablates MMSI4D with the BERT~\cite{devlin-etal-2019-bert} backbone by modifying one component of the full model at a time.
\new{For the two geometric inputs, namely the speaker motion (Sec.~\ref{sec:speaker_motion}) and relative positions (Eq.~\eqref{eq:relative_position}), we compute them from the projected 2D keypoints instead of removing them, so that the model still receives both cues.}
Head--candidate compatibility (Eq.~\eqref{eq:head_compat}) and residual refinement (Eq.~\eqref{eq:residual}) are removed entirely.
\new{Each of these modifications lowers average accuracy on both tasks.
The drops in the 2D settings, where the model still receives both cues, indicate that the benefit of each geometric input stems from its 3D form, in line with the results of Sec.~\ref{sec:effect_3d}.}

\paragraph{\new{Gaze and gesture cues.}}
\new{To examine the individual contributions of gaze and gesture cues, we split the speaker-motion joints into a gaze-related group (nose and eyes) and a gesture-related group (shoulders, elbows, and wrists), where the former serves as a proxy for head orientation.
As shown in Table~\ref{tab:headbody}, removing the features of either group reduces accuracy on every subset of both tasks, showing that both cues contribute to the performance of the full model.}

\begin{table}[t]
\centering
\caption{\textbf{3D gain across projective ambiguity.}
Accuracy (\%) of the 2D variant and MMSI4D (3D) across quartiles of $A$, pooled over both subsets and tasks.
$\Delta$ denotes the gain of 3D over 2D.}
\label{tab:when3d_ambiguity}
\scriptsize
\setlength{\tabcolsep}{5pt}
\begin{tabular*}{\columnwidth}{@{\extracolsep{\fill}}lcccc}
\toprule
Quartile & $A$ boundary & 2D & 3D & $\Delta$ \\
\midrule
$Q1$ (Low)      & $<0.48$        & \new{70.5} & 71.3 & \dpos{\new{+0.8}} \\
$Q2$ (Mid-Low)  & $0.48$--$0.76$ & \new{67.2} & 69.9 & \dpos{\new{+2.7}} \\
$Q3$ (Mid-High) & $0.76$--$1.01$ & \new{69.9} & 73.5 & \dpos{\new{+3.6}} \\
$Q4$ (High)     & $\ge 1.01$     & \new{65.8} & 70.5 & \dpos{\new{+4.7}} \\
\bottomrule
\end{tabular*}
\vspace{-2mm}
\end{table}

\subsection{When Does 3D Help?}
\label{sec:when_3d_helps}

\new{Sec.~\ref{sec:effect_3d} shows that 3D inputs improve accuracy over their 2D projections on average.
We now ask in which scenes this improvement arises.}
\new{Note that the 3D inputs preserve the spatial arrangement along the camera axis, which is key to disambiguating participants who appear adjacent in the image, as in Fig.~\ref{fig:motivation}.}
\new{We therefore expect the benefit of 3D to grow as such image-plane neighbors lie at increasingly different distances.
To test this, we quantify the difference for each scene and compare \new{BERT-based} MMSI4D with its 2D variant across scenes grouped accordingly.
}

\paragraph{\new{Quantifying projective ambiguity.}}

\new{For each test instance, \ie, a target utterance and its video window, we define the \emph{projective ambiguity} as $A=\operatorname{median}_i\lvert Z_i-Z_{\mathrm{nn}(i)}\rvert$}, where $Z_i$ is the camera-space $Z$ coordinate of participant $i$'s nose, taken as the median over the window, and $\mathrm{nn}(i)$ is the participant nearest to $i$ in the image plane.
\new{A larger $A$ indicates that image-plane neighbors differ more in their distances from the camera.}
Note that $A$ is computed solely from the reconstruction and is thus independent of any model prediction.
\new{Pooling both subsets and tasks, we divide the test instances for which $A$ is defined, 1,464 in total, into quartiles of $A$, from $Q1$ (lowest) to $Q4$ (highest).}

\begin{figure*}[ht!]
    \centering
    \makebox[0.5\linewidth][c]{\small Speaking Target Identification}%
    \makebox[0.5\linewidth][c]{\small Pronoun Coreference Resolution}\\[0.3em]
    \includegraphics[width=\linewidth]{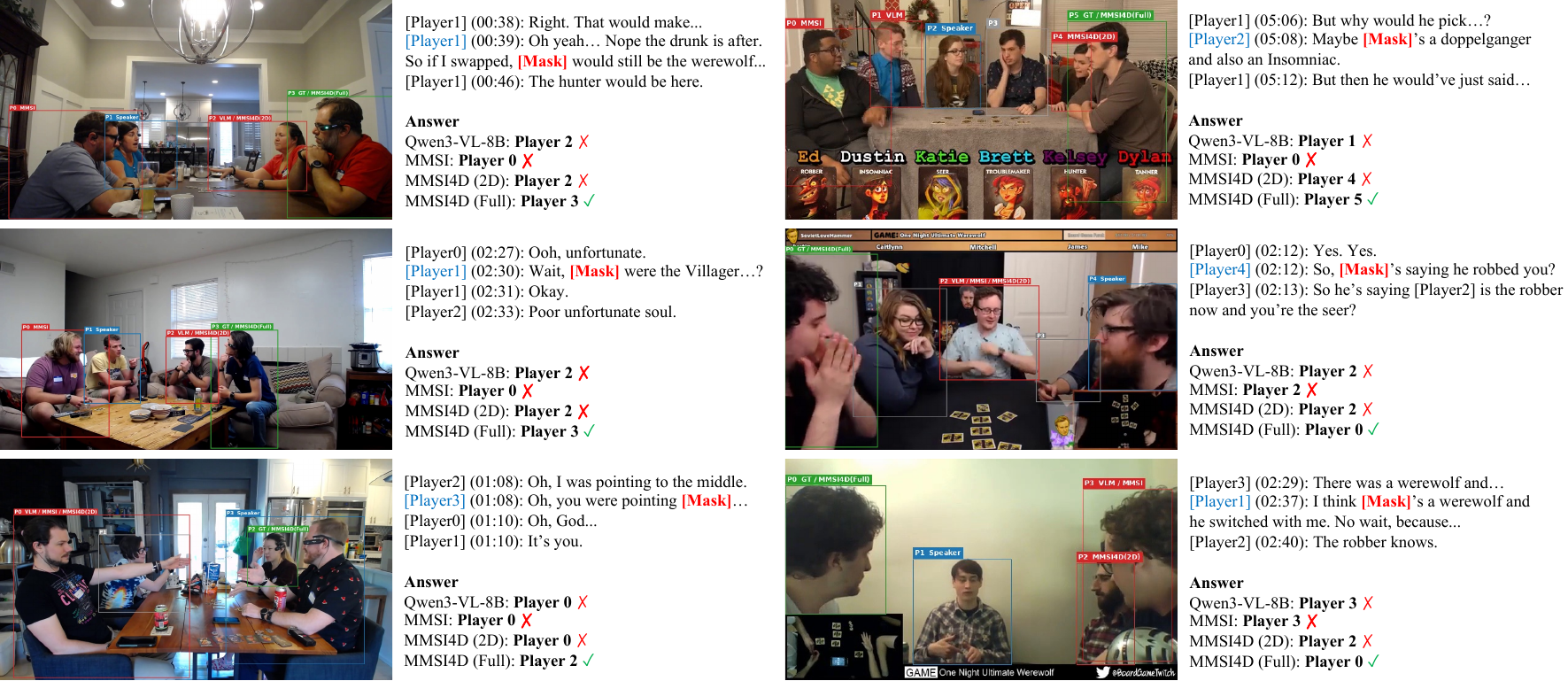}
    \caption{\textbf{Qualitative results.}
    Each panel shows an input frame and the dialogue context, where \textcolor{red}{\texttt{[MASK]}} marks the target reference.
    Boxes indicate the speaker (\textcolor{blue}{blue}), the ground truth (\textcolor{green!70!black}{green}), and the baseline predictions (\textcolor{red}{red}).
    \textcolor{green!70!black}{\checkmark} and \textcolor{red}{\xmark} denote correct and incorrect answers.}
    \label{fig:when3d_qual}
\end{figure*}

\paragraph{\new{Gain across ambiguity levels.}}

Consistent with our expectation, the gain of 3D \new{over 2D} increases steadily with $A$, as shown in Table~\ref{tab:when3d_ambiguity}.
In particular, 3D and 2D inputs yield comparable accuracy in $Q1$, where image-plane neighbors lie at similar distances from the camera.
Within each subset and task, $Q2$--$Q4$ likewise yield larger gains than $Q1$, as detailed in the supplementary material.
These results suggest that 3D becomes increasingly beneficial as these neighbors differ more in their distances from the camera, highlighting \new{the value of 3D} in resolving projective ambiguity.

\subsection{Qualitative Results}
\label{sec:qualitative}

\new{Fig.~\ref{fig:when3d_qual} compares the predictions of the full MMSI4D with those of Qwen3-VL-8B~\cite{bai2025qwen3}, MMSI~\cite{lee2024modeling}, and the 2D variant of MMSI4D.
All task-specific models use the BERT~\cite{devlin-etal-2019-bert} backbone.}
\new{The top-left example presents a concrete case of the projective ambiguity analyzed in Sec.~\ref{sec:when_3d_helps}.
Here, the 2D variant confuses the ground truth with another participant who lies in nearly the same direction from the speaker in the image.
In 3D, however, the ground truth sits closer to the camera, diagonally across from the speaker.
MMSI4D captures this difference along the camera axis and selects the ground truth.}
Additional examples with visualizations of the reconstructed arrangements are provided in the supplementary material.

\section{Conclusion}
\label{sec:conclusion}

We studied social referent reasoning in multi-party conversations, \new{a problem that requires} relating the speaker's verbal and nonverbal cues to the intended participant.
The nonverbal cues are directional and thus depend on the participants' 3D arrangement, yet existing models mostly locate participants in the image plane.
We therefore introduced MMSI4D, \new{which is, to our knowledge,} the first approach \new{to this problem} that explicitly encodes the speaker's 4D motion and \new{separately tokenizes each candidate's 3D relation to the speaker}.
\new{On two benchmark tasks, MMSI4D achieves more than twice the average gain} that \new{the prior 2D keypoint-based approach brings} over language-only models, and outperforms \new{zero-shot video LMMs with 7--8B parameters.}
Comparisons with the same 3D keypoints projected to 2D show that reasoning in 3D consistently contributes to this improvement, particularly where participants adjacent in the image lie at very different distances from the camera.

\paragraph{Limitations and future directions.}
\new{This work focused on the seated group conversations of the benchmark~\cite{lee2024modeling}, in which participants remain mostly stationary.
Extending our approach to unconstrained conversations, where participants interact while moving around freely, is a natural next step.
Adapting MMSI4D to online settings with streaming reasoning~\cite{li2025towards} and online HMR~\cite{Zhao_2026_CVPR}, together with automatic speaker identification~\cite{li2026omni}, is another promising direction toward real-world deployment.}
We hope this work serves as an initial step toward socially intelligent systems that reason about multimodal group interactions in physical 3D space.

\section*{LLM Usage Statement}
We used an LLM-based assistant (Claude, Anthropic) during the preparation of this paper.
The authors drafted the entire manuscript, and the assistant suggested edits to the text for clarity and grammar, \new{as well as to the visual presentation of tables and figures}.
All suggestions were reviewed and revised by the authors, who take full responsibility for the content.
No LLMs or generative AI tools were used to generate research ideas, experimental results, or \new{figure content}.
The video LMMs evaluated in Sec.~\ref{sec:setup} serve only as zero-shot baselines and were not used to prepare this paper.
\newpage
{
    \small
    \bibliographystyle{ieeenat_fullname}
    \bibliography{main}
}

\clearpage

\maketitlesupplementary

\setcounter{page}{1}

\appendix
\renewcommand{\thepage}{\arabic{page}}

\setcounter{table}{0}
\renewcommand{\thetable}{S\arabic{table}}
\setcounter{figure}{0}
\renewcommand{\thefigure}{S\arabic{figure}}

\noindent
This supplementary material provides additional information about the main paper.
Numbered sections, equations, tables, and figures refer to the main paper unless prefixed with ``S''.

\vspace{0.8em}
\noindent\textbf{Contents}
\vspace{0.4em}

\begingroup
\setlength{\parindent}{0pt}
\newcommand{\suppcontentsline}[3]{%
  \noindent\makebox[1.7em][l]{\textbf{#1}}%
  \hyperref[#2]{#3}\hfill\pageref{#2}\par
}
\suppcontentsline{A}{sec:supp_impl}{Implementation Details}
\suppcontentsline{B}{sec:supp_plane}{Effect of $XZ$-Plane Projection}
\suppcontentsline{C}{sec:supp_backbone}{\new{Results with Recent Language Backbones}}
\suppcontentsline{D}{sec:supp_persubset}{\new{Per-Subset Effect of 3D Inputs}}
\suppcontentsline{E}{sec:supp_when3d}{\new{Further Analysis of When 3D Helps}}
\suppcontentsline{F}{sec:supp_qual}{Additional Qualitative Results}
\suppcontentsline{G}{sec:supp_limitations}{Limitations}
\suppcontentsline{H}{sec:supp_impact}{Broader Impacts}
\endgroup
\section{Implementation Details}
\phantomsection
\label{sec:supp_impl}

\subsection{Data Preprocessing}
\label{sec:supp_data}

\paragraph{Reconstruction and tracking.}
\new{We reconstruct all participants in camera space with 4DHumans~\cite{goel2023humans}, which combines per-detection mesh recovery by HMR2.0 with 3D-aware tracking by PHALP~\cite{rajasegaran2022tracking}.}
The tracker matches people across frames by their reconstructed pose, location, and appearance.
\new{The reconstruction runs at the original video frame rate, whereas the keypoint annotations in the benchmark are provided at $5$\,Hz.
We therefore subsample each track to this rate, temporally aligning the visual inputs with the utterances.
We then link each track to the corresponding player identity in the dialogue annotations.
When a participant is temporarily undetected, we substitute their keypoints stored in a buffer, as in MMSI~\cite{lee2024modeling}.}
We will make these reconstructions and identity links publicly available to facilitate future research.

\paragraph{Camera model.}
\new{4DHumans~\cite{goel2023humans} assumes a fixed focal length of $5000$ pixels for a $256\times256$ input crop, which corresponds to a nearly orthographic camera with a field of view of about $3^\circ$.
While this assumption is sufficient for reconstructing each person individually, it places participants more than an order of magnitude farther along the optical axis than a typical perspective camera.
To obtain relations at a realistic scale, we use nominal intrinsics derived from the image size, as is common in human mesh recovery (HMR)~\cite{li2022cliff,shin2024wham}.
Specifically, we set the focal length to the longer image side and the principal point to the image center, which gives a horizontal field of view of about $53^\circ$.
This change amounts to rescaling each person's distance along the optical axis by the constant factor $256/5000$ and leaves the body poses unchanged.
Note that the 2D variant in Sec.~4.3 takes the 3D keypoints projected with these intrinsics, so that it shares the same camera as MMSI4D.}

\begin{table}[t!]
\centering
\caption{\textbf{Module configurations of MMSI4D.}
Numbers in parentheses denote the layer widths from input to output, with $512^{4}$ indicating four $512$-dimensional layers.
For the Transformers, they denote the numbers of layers and heads, and the FFN dimension.}
\label{tab:supp_arch}
\footnotesize
\setlength{\tabcolsep}{4pt}
\renewcommand{\arraystretch}{1.15}
\begin{tabular*}{\columnwidth}{@{\extracolsep{\fill}}lll@{}}
\toprule
Module & Role & Configuration \\
\midrule
\multicolumn{3}{@{}l}{\textit{Speaker motion} (Sec.~3.3)} \\
$E_{\mathrm{point}}$ & per-joint point encoder & MLP $(3,\,64^{3})$ \\
$E_{\mathrm{motion}}$ & per-frame motion encoder & MLP $(J{\cdot}64,\,512^{4})$ \\
PE & temporal positional encoding & sinusoidal, $T'{=}8$ \\
\midrule
\multicolumn{3}{@{}l}{\textit{Candidate-wise tokenization} (Sec.~3.5)} \\
$E_{\mathrm{rel}}$ & relation encoder & MLP $(7,\,64,\,512)$ \\
$E_{\mathrm{vis}}$ & visual Transformer & $(3,\,8,\,1024)$ \\
$\mathbf{e}_i$ & slot embedding & Embedding $(C,\,512)$ \\
\midrule
\multicolumn{3}{@{}l}{\textit{Multimodal fusion} (Sec.~3.6)} \\
Backbone & pretrained language backbone & $768$-d hidden state \\
$E_{\mathrm{lang}}$ & language encoder & Linear $(768,\,512)$ \\
$E_{\mathrm{multi}}$ & multimodal Transformer & $(2,\,8,\,1024)$ \\
$\mathbf{h}^{\mathrm{agg}}$ & aggregation token & $(1,\,512)$ \\
\midrule
\multicolumn{3}{@{}l}{\textit{Prediction} (Sec.~3.6)} \\
Classifier & global logits & Linear $(512,\,C)$ \\
$f_{\mathrm{res}}$ & residual scorer & Linear $(512,\,1)$ \\
\bottomrule
\end{tabular*}
\vspace{-3mm}
\end{table}

\subsection{Network Architecture}
\label{sec:supp_arch}
\new{In addition to the hidden size and numbers of Transformer layers in the main paper, Table~\ref{tab:supp_arch} lists the configuration of every module.
Here, $J{=}9$ and $C{=}6$ denote the numbers of speaker-motion joints and candidate slots, respectively.}

\subsection{Training and Inference Details}
\label{sec:supp_training}

\new{In this section, we describe how participants are assigned to slots and how the augmentations of MMSI~\cite{lee2024modeling} are applied to our 3D inputs.}

\paragraph{Slot handling.}
\new{Each participant, including the speaker, occupies one of the $C$ slots.
For the speaker's own slot, the relative position $\mathbf{r}_i$, direction $\mathbf{u}_i$, and distance $d_i$ are zero by definition.
The head--candidate compatibility $c_i$ does not apply to this slot and is set to zero, so the speaker is identified by the indicator $s_i$.
When $c_i$ cannot be computed within the visual window, we likewise set it to zero.
Unused slots are masked in $E_{\mathrm{vis}}$ and $E_{\mathrm{multi}}$ and excluded from the residual refinement.}

\paragraph{Mirroring augmentation.}
We apply the mirroring described in \new{Sec.~3.7} to each training sample with a probability of $0.5$.
\new{This operation negates the camera $X$ coordinates of the speaker motion and of each candidate's relative position and direction, and swaps the left and right joints to keep the joint indices consistent with the mirrored body.}
The dialogue and the target labels remain unchanged.

\paragraph{Permutation augmentation.}
\new{We apply the slot shuffling described in Sec.~3.7 to every sample during training only.
The player identifiers in the dialogue, the target labels, and the per-participant visual inputs are remapped accordingly, while unused slots keep their positions.}

\begin{table}[t]
\centering
\caption{\textbf{Effect of $XZ$-plane projection.}
Accuracy (\%) of MMSI4D with BERT~\cite{devlin-etal-2019-bert}.
$\Delta$ denotes $XZ$ minus $XYZ$.}
\label{tab:supp_plane}
\scriptsize
\setlength{\tabcolsep}{3.5pt}
\renewcommand{\arraystretch}{1.12}
\begin{tabular*}{\columnwidth}{@{\extracolsep{\fill}}l ccc ccc}
\toprule
& \multicolumn{3}{c}{STI} & \multicolumn{3}{c}{PCR} \\
\cmidrule(lr){2-4}\cmidrule(lr){5-7}
Coordinates & YouTube & Ego4D & Avg. & YouTube & Ego4D & Avg. \\
\midrule
$XYZ$ & 72.5 & 71.0 & 71.8 & \textbf{69.6} & 51.8 & 60.7 \\
$XZ$ (Ours) & \textbf{74.2} & \textbf{75.6} & \textbf{74.9} & 69.0 & \textbf{58.9} & \textbf{64.0} \\
\midrule
$\Delta$ & +1.7 & +4.6 & \dpos{+3.1} & \textminus0.6 & +7.1 & \dpos{+3.3} \\
\bottomrule
\end{tabular*}
\vspace{-2mm}
\end{table}

\subsection{Zero-shot Video LMM Prompts}
\label{sec:supp_prompt}

The task-specific models require a \texttt{[MASK]} token in the dialogue to mark the referent to be predicted.
This token is appended to the target utterance as \texttt{(To [MASK])} for Speaking Target Identification (STI) and replaces the target pronoun for Pronoun Coreference Resolution (PCR).
In Figs.~2 and~3, we show it in place of the target reference for clarity.
The four video LMMs in Table~1~\cite{zhang2024llava,Qwen2.5-VL,wang2025internvl3_5,bai2025qwen3} are instead evaluated with a single prompt per task, shared across models.
Each prompt contains the dialogue context, the candidate list, and eight frames in chronological order, in which each visible participant is labeled with an ID such as P0.
In these prompts, we use \texttt{[TARGET]} to mark the target utterance and \texttt{[PRONOUN]} to represent the pronoun for PCR.
For STI, we use the utterance as is, since the second-person expression does not indicate the referent and is likewise kept for the task-specific models.
Each LMM is asked to return a single candidate ID from the list.
Figures~\ref{fig:supp_lmm_input_sti} and~\ref{fig:supp_lmm_input_pcr} show an example input for STI and PCR, respectively, and the boxes below give the corresponding prompt templates.

\par\vspace{5pt}
\noindent\begin{minipage}{\columnwidth}
\begin{promptquote}
(*@\rmfamily\bfseries Prompt template for STI@*)

You are solving a multi-party social interaction benchmark. Eight visual frames are shown in chronological order. Visible participants are labeled with participant IDs. Use only the provided visual frames and conversation.

Task: Speaking Target Identification (STI)
Determine which participant the speaker of the TARGET utterance is directly speaking to.

Candidate IDs: {candidate_ids}
Conversation: {conversation}
TARGET utterance: {target_utterance}

Which participant is the speaker directly addressing?
Return exactly one candidate ID from the candidate list.
Do not provide an explanation.
\end{promptquote}
\end{minipage}
\par\vspace{7pt}

\noindent\begin{minipage}{\columnwidth}
\begin{promptquote}
(*@\rmfamily\bfseries Prompt template for PCR@*)

You are solving a multi-party social interaction benchmark. Eight visual frames are shown in chronological order. Visible participants are labeled with participant IDs. Use only the provided visual frames and conversation.

Task: Pronoun Coreference Resolution (PCR)
Determine which participant the marked pronoun in the TARGET utterance refers to.

Candidate IDs: {candidate_ids}
Conversation: {conversation}
TARGET utterance: {target_utterance}

Which participant does the marked pronoun refer to?
Return exactly one candidate ID from the candidate list.
Do not provide an explanation.
\end{promptquote}
\end{minipage}

\section{Effect of $XZ$-Plane Projection}
\phantomsection
\label{sec:supp_plane}

\new{As defined in Eq.~(2), we compute the relative position $\mathbf{r}_i$ of each candidate with respect to the speaker on the camera-$XZ$ plane, similar to the horizontal-plane placement used in multi-person motion generation~\cite{yu2026socialgen}.
This choice is motivated by the observation that participants in the benchmark mostly remain seated at similar heights, so their arrangement varies mainly within this plane.
Table~\ref{tab:supp_plane} compares it with a variant that retains all three $XYZ$ coordinates, using the BERT backbone~\cite{devlin-etal-2019-bert}.
The projection yields higher average accuracy on both tasks, with the gains coming mainly from Ego4D.
This suggests that the projection reduces noise from the vertical coordinate, allowing the model to learn the spatial arrangement more effectively.}

\begin{table*}[t!]
\centering
\caption{\textbf{Results with recent BERT-style backbones} on Speaking Target Identification (STI) and Pronoun Coreference Resolution (PCR) on the YouTube and Ego4D subsets.
We report accuracy (\%), with Avg.\ denoting the unweighted mean of the two subsets, computed before rounding.
We retrain all task-specific methods with the same data splits and evaluation protocol as in Table~1.
\textbf{Bold} and \underline{underline} indicate the best and second-best results within each language-backbone group.}
\label{tab:supp_recent_backbone}

\scriptsize
\setlength{\tabcolsep}{3.3pt}

\resizebox{\textwidth}{!}{
\begin{tabular}{l cccc ccc ccc}
\toprule
& \multicolumn{4}{c}{\textbf{Representation / Modeling}}
& \multicolumn{3}{c}{\textbf{Speaking Target Identification}}
& \multicolumn{3}{c}{\textbf{Pronoun Coreference Resolution}} \\
\cmidrule(lr){2-5} \cmidrule(lr){6-8} \cmidrule(lr){9-11}
\textbf{Method} &
\shortstack{\textbf{Aligned?}} & \textbf{Explicit?} & \textbf{3D?} & \shortstack{\textbf{Candidate-wise?}} &
\textbf{YouTube} & \textbf{Ego4D} & \textbf{Avg.} &
\textbf{YouTube} & \textbf{Ego4D} & \textbf{Avg.} \\
\midrule
DeBERTa-v3~\cite{he2021debertav3} & \textcolor{red}{\xmark} & \textcolor{red}{\xmark} & \textcolor{red}{\xmark} & \textcolor{red}{\xmark}
& 74.5 & 64.2 & 69.4 & 71.7 & 46.4 & 59.1 \\
DeBERTa-v3 + DINOv2~\cite{oquab2023dinov2} & \textcolor{red}{\xmark} & \textcolor{red}{\xmark} & \textcolor{red}{\xmark} & \textcolor{red}{\xmark}
& \underline{75.3} & 60.2 & 67.7 & 74.0 & 43.8 & 58.9 \\
DeBERTa-v3 + MViT~\cite{fan2021multiscale,lai2023werewolf} & \textcolor{red}{\xmark} & \textcolor{red}{\xmark} & \textcolor{red}{\xmark} & \textcolor{red}{\xmark}
& 74.5 & 62.0 & 68.3 & \underline{75.0} & 44.8 & 59.9 \\
DeBERTa-v3-based MMSI~\cite{lee2024modeling} & \textcolor{green!70!black}{\checkmark} & \textcolor{green!70!black}{\checkmark} & \textcolor{red}{\xmark} & \textcolor{red}{\xmark}
& 74.1 & \underline{66.1} & \underline{70.1} & 73.4 & \underline{49.1} & \underline{61.3} \\
\textbf{DeBERTa-v3-based MMSI4D (Ours)} & \textcolor{green!70!black}{\checkmark} & \textcolor{green!70!black}{\checkmark} & \textcolor{green!70!black}{\checkmark} & \textcolor{green!70!black}{\checkmark}
& \textbf{77.1} & \textbf{76.7} & \textbf{76.9} & \textbf{75.1} & \textbf{53.0} & \textbf{64.1} \\
\cmidrule(lr){1-11}
ModernBERT~\cite{warner2025smarter} & \textcolor{red}{\xmark} & \textcolor{red}{\xmark} & \textcolor{red}{\xmark} & \textcolor{red}{\xmark}
& 74.2 & 69.9 & 72.0 & 70.4 & 49.1 & 59.7 \\
ModernBERT + DINOv2~\cite{oquab2023dinov2} & \textcolor{red}{\xmark} & \textcolor{red}{\xmark} & \textcolor{red}{\xmark} & \textcolor{red}{\xmark}
& 72.1 & 70.5 & 71.3 & 72.8 & 48.2 & 60.5 \\
ModernBERT + MViT~\cite{fan2021multiscale,lai2023werewolf} & \textcolor{red}{\xmark} & \textcolor{red}{\xmark} & \textcolor{red}{\xmark} & \textcolor{red}{\xmark}
& 72.5 & \underline{76.0} & \underline{74.2} & \underline{74.4} & 43.7 & 59.0 \\
ModernBERT-based MMSI~\cite{lee2024modeling} & \textcolor{green!70!black}{\checkmark} & \textcolor{green!70!black}{\checkmark} & \textcolor{red}{\xmark} & \textcolor{red}{\xmark}
& \underline{75.1} & 70.0 & 72.6 & 72.7 & \underline{50.9} & \underline{61.8} \\
\textbf{ModernBERT-based MMSI4D (Ours)} & \textcolor{green!70!black}{\checkmark} & \textcolor{green!70!black}{\checkmark} & \textcolor{green!70!black}{\checkmark} & \textcolor{green!70!black}{\checkmark}
& \textbf{75.6} & \textbf{76.1} & \textbf{75.9} & \textbf{74.8} & \textbf{57.1} & \textbf{66.0} \\
\bottomrule
\end{tabular}
}
\vspace{-2.5mm}
\end{table*}

\begin{table}[t]
\centering
\caption{\textbf{Per-subset breakdown of Table~2.}
Accuracy (\%) of MMSI4D on each subset, with 3D keypoints (3D) and with their 2D projections (2D). $\Delta$~denotes the gain of 3D over 2D.}
\label{tab:supp_repr_persubset}
\footnotesize
\setlength{\tabcolsep}{3.5pt}
\renewcommand{\arraystretch}{1.12}
\begin{tabular*}{\columnwidth}{@{\extracolsep{\fill}}l ccc ccc}
\toprule
& \multicolumn{3}{c}{YouTube} & \multicolumn{3}{c}{Ego4D} \\
\cmidrule(lr){2-4}\cmidrule(lr){5-7}
Backbone & 2D & 3D & $\Delta$ & 2D & 3D & $\Delta$ \\
\midrule
\multicolumn{7}{l}{\textit{Speaking Target Identification (STI)}} \\
BERT~\cite{devlin-etal-2019-bert}  & 71.8 & 74.2 & \dpos{+2.4} & 69.3 & 75.6 & \dpos{+6.3} \\
RoBERTa~\cite{liu2019roberta}      & 73.6 & 74.7 & \dpos{+1.1} & 72.2 & 76.7 & \dpos{+4.5} \\
ELECTRA~\cite{clark2020electra}    & 70.8 & 72.1 & \dpos{+1.3} & 65.3 & 71.6 & \dpos{+6.3} \\
\midrule
\multicolumn{7}{l}{\textit{Pronoun Coreference Resolution (PCR)}} \\
BERT~\cite{devlin-etal-2019-bert}  & 66.9 & 69.0 & \dpos{+2.1} & 54.5 & 58.9 & \dpos{+4.4} \\
RoBERTa~\cite{liu2019roberta}      & 73.2 & 75.0 & \dpos{+1.8} & 56.3 & 60.7 & \dpos{+4.4} \\
ELECTRA~\cite{clark2020electra}    & 69.0 & 69.4 & \dpos{+0.4} & 50.9 & 54.5 & \dpos{+3.6} \\
\bottomrule
\end{tabular*}
\vspace{-2mm}
\end{table}

\section{\new{Results with Recent Language Backbones}}
\phantomsection
\label{sec:supp_backbone}

\new{Beyond the three language backbones in the main paper, we examine whether the advantage of MMSI4D generalizes to more recent language models.
To this end, we additionally train all task-specific methods with DeBERTa-v3-base~\cite{he2021debertav3} and ModernBERT-base~\cite{warner2025smarter}.
For MMSI~\cite{lee2024modeling}, we adapt its official code to the two backbones.
For the baselines with holistic visual features, we use the MViT~\cite{fan2021multiscale} features released by \citet{lai2023werewolf} and extract DINOv2~\cite{oquab2023dinov2} features as described in MMSI.
Table~\ref{tab:supp_recent_backbone} shows that MMSI4D achieves the best average accuracy on both tasks with both backbones.
Together with Table~1, this indicates that the advantage of MMSI4D generalizes across all five backbones.}

\section{\new{Per-Subset Effect of 3D Inputs}}
\phantomsection
\label{sec:supp_persubset}

\new{Table~\ref{tab:supp_repr_persubset} breaks down the MMSI4D results in Table~2 by subset.
MMSI4D outperforms its 2D variant on both subsets for every backbone and task, with a consistently larger gain on Ego4D.
Note that Ego4D instances tend to have higher projective ambiguity $A$ than YouTube instances, as analyzed in Sec.~\ref{sec:supp_when3d}.
Together with the trend in Sec.~4.6, where the gain of 3D increases with $A$, these results support the explanation in Sec.~4.2 for the larger gains of MMSI4D on Ego4D.}

\section{\new{Further Analysis of When 3D Helps}}
\phantomsection
\label{sec:supp_when3d}

We extend the analysis in \new{Sec.~4.6} to each subset and task, using the BERT-based MMSI4D and its 2D variant from Table~6.

\begin{figure}[t]
\centering
\definecolor{cYouTube}{RGB}{22,78,66}
\definecolor{cEgo}{HTML}{D66E1A}
\begin{tikzpicture}
\begin{axis}[
  width=\columnwidth,
  height=1.65in,
  xmin=0,
  xmax=2.5,
  ymin=0,
  ymax=1.80,
  xlabel={Projective ambiguity $A$},
  ylabel={Density},
  xtick={0,0.5,1.0,1.5,2.0,2.5},
  ytick={0,0.5,1.0,1.5},
  ymajorgrids=true,
  grid style={gray!12, line width=0.35pt},
  axis x line*=bottom,
  axis y line*=left,
  axis line style={black!75},
  tick style={black!75},
  tick label style={font=\scriptsize, color=black!90},
  label style={font=\small},
  xlabel style={at={(ticklabel cs:0.5)}, anchor=north, yshift=-1.0ex},
  legend style={
    at={(0.5,1.0)},
    anchor=south,
    yshift=5pt,
    inner sep=1pt,
    legend columns=2,
    draw=none,
    fill=none,
    font=\scriptsize,
    /tikz/every even column/.append style={column sep=1.6em},
  },
  clip=false,
  enlarge x limits=false,
]
\addlegendimage{area legend, fill=cYouTube, fill opacity=0.18, draw=cYouTube, line width=0.5pt}
\addlegendentry{YouTube}
\addlegendimage{area legend, fill=cEgo, fill opacity=0.22, draw=cEgo, line width=0.5pt}
\addlegendentry{Ego4D}
\addplot[forget plot, fill=cYouTube, fill opacity=0.18, draw=cYouTube, line width=0.5pt]
  coordinates {
    (0.0000,0) (0.0000,0.1633) (0.0625,0.1633) (0.0625,0.4218)
    (0.1250,0.4218) (0.1250,0.5170) (0.1875,0.5170) (0.1875,0.6939)
    (0.2500,0.6939) (0.2500,0.7211) (0.3125,0.7211) (0.3125,0.9524)
    (0.3750,0.9524) (0.3750,0.7211) (0.4375,0.7211) (0.4375,0.7347)
    (0.5000,0.7347) (0.5000,1.1429) (0.5625,1.1429) (0.5625,1.0340)
    (0.6250,1.0340) (0.6250,0.9388) (0.6875,0.9388) (0.6875,1.2381)
    (0.7500,1.2381) (0.7500,1.3469) (0.8125,1.3469) (0.8125,1.2925)
    (0.8750,1.2925) (0.8750,0.7619) (0.9375,0.7619) (0.9375,1.0204)
    (1.0000,1.0204) (1.0000,0.6395) (1.0625,0.6395) (1.0625,0.3946)
    (1.1250,0.3946) (1.1250,0.1633) (1.1875,0.1633) (1.1875,0.3129)
    (1.2500,0.3129) (1.2500,0.0544) (1.3125,0.0544) (1.3125,0.0000)
    (1.3750,0.0000) (1.3750,0.1633) (1.4375,0.1633) (1.4375,0.0136)
    (1.5000,0.0136) (1.5000,0.0544) (1.5625,0.0544) (1.5625,0.0136)
    (1.6250,0.0136) (1.6250,0.0136) (1.6875,0.0136) (1.6875,0.0000)
    (1.7500,0.0000) (1.7500,0.0000) (1.8125,0.0000) (1.8125,0.0408)
    (1.8750,0.0408) (1.8750,0.0136) (1.9375,0.0136) (1.9375,0.0000)
    (2.0000,0.0000) (2.0000,0.0136) (2.0625,0.0136) (2.0625,0.0272)
    (2.1250,0.0272) (2.1250,0.0272) (2.1875,0.0272) (2.1875,0.0000)
    (2.2500,0.0000) (2.2500,0.0000) (2.3125,0.0000) (2.3125,0.0000)
    (2.3750,0.0000) (2.3750,0.0272) (2.4375,0.0272) (2.4375,0.0000)
    (2.5000,0.0000) (2.5000,0)
  } \closedcycle;
\addplot[forget plot, fill=cEgo, fill opacity=0.22, draw=cEgo, line width=0.5pt]
  coordinates {
    (0.0000,0) (0.0000,0.2778) (0.0625,0.2778) (0.0625,0.2222)
    (0.1250,0.2222) (0.1250,0.0556) (0.1875,0.0556) (0.1875,0.0556)
    (0.2500,0.0556) (0.2500,0.2222) (0.3125,0.2222) (0.3125,0.1667)
    (0.3750,0.1667) (0.3750,0.1667) (0.4375,0.1667) (0.4375,0.1111)
    (0.5000,0.1111) (0.5000,0.1111) (0.5625,0.1111) (0.5625,0.1111)
    (0.6250,0.1111) (0.6250,0.2222) (0.6875,0.2222) (0.6875,0.3333)
    (0.7500,0.3333) (0.7500,0.5000) (0.8125,0.5000) (0.8125,0.1111)
    (0.8750,0.1111) (0.8750,1.0000) (0.9375,1.0000) (0.9375,0.6667)
    (1.0000,0.6667) (1.0000,1.1111) (1.0625,1.1111) (1.0625,0.5556)
    (1.1250,0.5556) (1.1250,0.3889) (1.1875,0.3889) (1.1875,0.7222)
    (1.2500,0.7222) (1.2500,1.0000) (1.3125,1.0000) (1.3125,0.0556)
    (1.3750,0.0556) (1.3750,0.2778) (1.4375,0.2778) (1.4375,0.8889)
    (1.5000,0.8889) (1.5000,0.7778) (1.5625,0.7778) (1.5625,0.3889)
    (1.6250,0.3889) (1.6250,1.0000) (1.6875,1.0000) (1.6875,0.5556)
    (1.7500,0.5556) (1.7500,0.7778) (1.8125,0.7778) (1.8125,0.6667)
    (1.8750,0.6667) (1.8750,1.0000) (1.9375,1.0000) (1.9375,0.5556)
    (2.0000,0.5556) (2.0000,0.4444) (2.0625,0.4444) (2.0625,0.2222)
    (2.1250,0.2222) (2.1250,0.0556) (2.1875,0.0556) (2.1875,0.0556)
    (2.2500,0.0556) (2.2500,0.0556) (2.3125,0.0556) (2.3125,0.1111)
    (2.3750,0.1111) (2.3750,0.0000) (2.4375,0.0000) (2.4375,0.0000)
    (2.5000,0.0000) (2.5000,0)
  } \closedcycle;
\end{axis}
\end{tikzpicture}
\caption{\textbf{Per-subset distribution of projective ambiguity.}
Histograms of $A$ for both subsets, each normalized to unit area, with the horizontal axis truncated at $2.5$.}
\label{fig:supp_subset_amb}
\vspace{-2mm}
\end{figure}
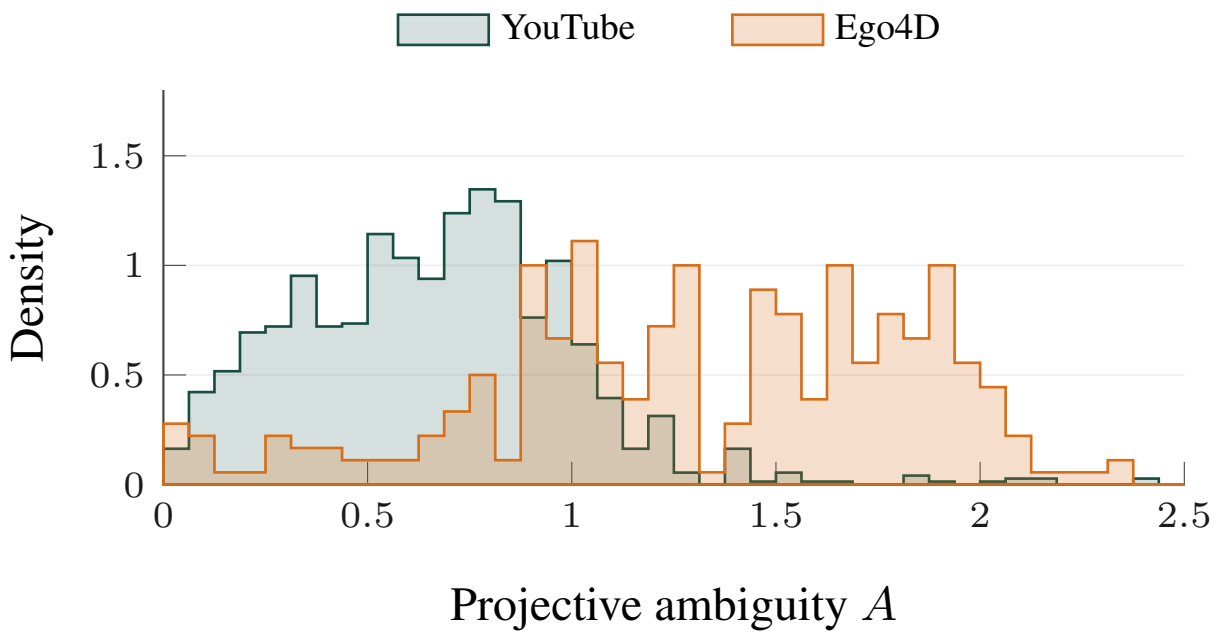

\paragraph{Per-subset distribution of projective ambiguity.}
\new{We first compare the distribution of projective ambiguity $A$ between the two subsets.
Recall that $A$ measures how much image-plane neighbors differ in their distances from the camera.
Figure~\ref{fig:supp_subset_amb} shows that this distribution is shifted toward higher values for Ego4D, with a median of $1.30$ compared with $0.68$ for YouTube.
Consequently, Ego4D instances make up the majority of the highest-ambiguity quartile $Q4$ of Table~6, whereas $Q1$--$Q3$ contain mainly YouTube instances.}

\paragraph{Per-subset 3D gain across ambiguity.}
\new{Table~\ref{tab:when3d_dataset} splits the test instances of Table~6 by subset, pooling both tasks.
We apply the pooled quartile boundaries to each subset, so the instance count $n$ in each quartile follows the distributions of $A$ above.
The gains on Ego4D vary more across quartiles, likely due to its small number of instances in $Q1$--$Q3$.
Nevertheless, $Q2$--$Q4$ yield larger gains than $Q1$ on both subsets, consistent with the trend in Sec.~4.6.}

\paragraph{Per-task 3D gain across ambiguity.}
\new{Table~\ref{tab:supp_quartile} instead splits the same instances by task, pooling both subsets.
For both tasks, $Q2$--$Q4$ also yield larger gains than $Q1$, and the overall gain is higher for STI.
We hypothesize that this is because second-person expressions such as ``you'' are often accompanied by gaze or gestures directed at the addressee.}

\paragraph{\new{Decomposing the 3D gain.}}
\new{We next take a closer look at which instances account for the gains in Table~6.
To this end, we count the instances classified correctly only by MMSI4D (\emph{positive flips}) and only by the 2D variant (\emph{negative flips}).
Figure~\ref{fig:rescue_harm} shows that positive flips increase steadily with $A$, from $6.3\%$ in $Q1$ to $13.4\%$ in $Q4$, whereas negative flips remain at $5.5\%$ in $Q1$--$Q3$ and rise only to $8.7\%$ in $Q4$.
The gap between the two curves corresponds to the 3D gain in Table~6 and widens as $A$ increases.
An exact McNemar test~\cite{mcnemar1947note} on these flips further confirms that the gain over all test instances is statistically significant, with $p=0.005$.}

\begin{table}[t!]
\centering
\newcommand{\dw}[1]{\makebox[2.4em][c]{#1}}   %
\caption{\textbf{3D gain by subset.}
Accuracy (\%) with 2D and 3D inputs in each quartile of $A$, using the BERT~\cite{devlin-etal-2019-bert} backbone.
$\Delta$ and $n$ denote the 3D gain and the instance count, respectively.}
\label{tab:when3d_dataset}
\footnotesize
\setlength{\tabcolsep}{3.5pt}
\renewcommand{\arraystretch}{1.1}
\begin{tabular*}{\columnwidth}{@{\extracolsep{\fill}}l l cccc c}
\toprule
Subset & & $Q1$ & $Q2$ & $Q3$ & $Q4$ & All \\
\midrule
\multirow{3}{*}{YouTube}
 & 2D & 71.1 & 67.9 & 71.6 & 65.4 & 69.6 \\
 & 3D & 72.3 & 70.8 & 73.5 & 69.9 & 71.9 \\
 & $\Delta$ & \dpos{\dw{+1.2}} & \dpos{\dw{+2.9}} & \dpos{\dw{+1.9}} & \dpos{\dw{+4.5}} & \dpos{\dw{+2.3}} \\
\cmidrule(lr){2-7}
 & $n$ & 343 & 349 & 328 & 156 & 1176 \\
\midrule
\multirow{3}{*}{Ego4D}
 & 2D & 60.9 & 52.9 & 55.3 & 66.2 & 63.5 \\
 & 3D & 56.5 & 52.9 & 73.7 & 71.0 & 69.1 \\
 & $\Delta$ & \dneg{\dw{\textminus4.4}} & \dzero{\dw{0.0}} & \dpos{\dw{+18.4}} & \dpos{\dw{+4.8}} & \dpos{\dw{+5.6}} \\
\cmidrule(lr){2-7}
 & $n$ & 23 & 17 & 38 & 210 & 288 \\
\bottomrule
\end{tabular*}
\vspace{-1mm}
\end{table}
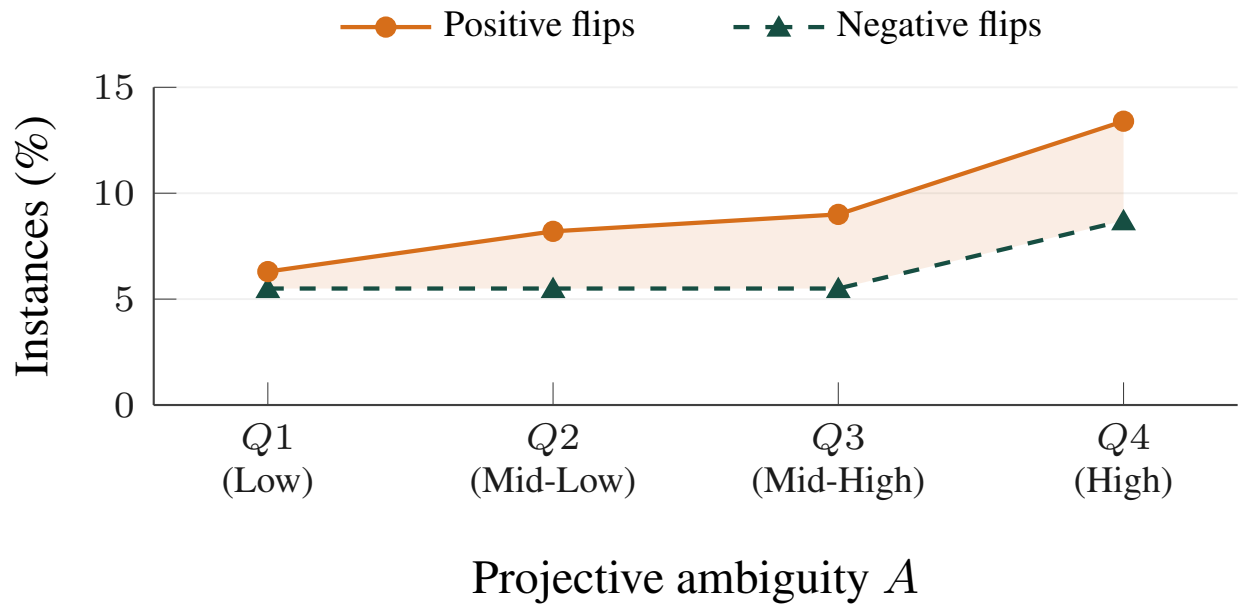
\begin{figure}[t]
\centering
\definecolor{cFlipPos}{HTML}{D66E1A}
\definecolor{cFlipNeg}{RGB}{22,78,66}
\begin{tikzpicture}
\begin{axis}[
  width=1.05\columnwidth,  height=1.45in,
  xmin=0.6,
  xmax=4.4,
  ymin=0,
  ymax=15,
  xlabel={Projective ambiguity $A$},
  ylabel={Instances (\%)},
  xtick={1,2,3,4},
  xticklabels={{$Q1$\\ (Low)}, {$Q2$\\ (Mid-Low)}, {$Q3$\\ (Mid-High)}, {$Q4$\\ (High)}},
  ytick={0,5,10,15},
  ymajorgrids=true,
  grid style={gray!12, line width=0.35pt},
  axis x line*=bottom,
  axis y line*=left,
  axis line style={black!75},
  tick style={black!75},
  tick label style={font=\scriptsize, color=black!90},
  x tick label style={align=center, font=\scriptsize, color=black!90},
  label style={font=\small},
  xlabel style={at={(ticklabel cs:0.5)}, anchor=north, yshift=-1.0ex},
  legend style={
    at={(0.5,1.0)},
    anchor=south,
    yshift=5pt,
    inner sep=1pt,
    legend columns=2,
    draw=none,
    fill=none,
    font=\scriptsize,
    /tikz/every even column/.append style={column sep=1.6em},
  },
  clip=false,
  enlarge x limits=false,
  reverse legend,
]
\addplot[forget plot, draw=none, fill=cFlipPos, fill opacity=0.12]
  coordinates {(1,6.3) (2,8.2) (3,9.0) (4,13.4) (4,8.7) (3,5.5) (2,5.5) (1,5.5)} -- cycle;
\addplot[cFlipNeg, line width=0.8pt, dashed, mark=triangle*, mark size=1.9pt,
    mark options={solid, fill=cFlipNeg, yshift=-0.45pt},
    legend image code/.code={
      \draw[cFlipNeg, line width=0.8pt, dashed] (0cm,0cm) -- (0.3cm,0cm);
      \draw[cFlipNeg, line width=0.8pt, dashed] (0.6cm,0cm) -- (0.3cm,0cm);
      \draw[cFlipNeg, only marks, mark=triangle*, mark size=1.9pt,
        mark options={solid, fill=cFlipNeg, yshift=-0.45pt}] plot coordinates {(0.3cm,0cm)};
    }]
  coordinates {(1,5.5) (2,5.5) (3,5.5) (4,8.7)};
\addlegendentry{Negative flips}
\addplot[cFlipPos, line width=0.8pt, mark=*, mark size=1.6pt, mark options={solid, fill=cFlipPos}]
  coordinates {(1,6.3) (2,8.2) (3,9.0) (4,13.4)};
\addlegendentry{Positive flips}
\end{axis}
\end{tikzpicture}
\caption{\textbf{Prediction flips across quartiles.}
Percentage of instances per quartile classified correctly only by MMSI4D (positive) or only by the 2D variant (negative), using the BERT~\cite{devlin-etal-2019-bert} backbone.}
\label{fig:rescue_harm}
\vspace{-2mm}
\end{figure}

\paragraph{Comparison with MMSI across ambiguity.}
\new{Figure~\ref{fig:supp_quartile_mmsi} adds the released BERT-based MMSI~\cite{lee2024modeling} to the comparison in Table~6.
Note that MMSI uses 2D keypoints from AlphaPose~\cite{fang2022alphapose}, whereas the 2D variant takes projected HMR2.0 keypoints.
The gain of MMSI4D over MMSI grows with $A$, from $0.5$ points in $Q1$ to $9.0$ points in $Q4$, and is statistically significant over all test instances under the exact McNemar test, with $p<0.001$.
MMSI and the 2D variant, both of which operate on the image plane, perform comparably in $Q1$--$Q3$, with the 2D variant higher in $Q4$.
In contrast, MMSI4D remains more accurate than both in every quartile, including the high-ambiguity scenes of $Q4$.}

\paragraph{\new{Removing the 3D arrangement at test time.}}
\new{We further test whether MMSI4D makes use of the 3D arrangement during inference.
To this end, we evaluate the trained BERT-based MMSI4D of Table~2 on flattened 3D inputs, in which all participants are placed at the same distance from the camera.
Specifically, for each frame, we translate every participant along their viewing ray to their average distance, while keeping their positions in the image unchanged.
Whereas the 2D variant is \new{separately} trained on 2D inputs, this test keeps the trained MMSI4D fixed and only equalizes the participants' distances along the optical axis in the test instances.
Table~\ref{tab:flatten} shows that this change lowers accuracy on both tasks to a level close to that of the 2D variant.
This confirms that MMSI4D makes use of the 3D arrangement of participants during inference.}

\begin{table}[t]
\centering
\newcommand{\dw}[1]{\makebox[2.4em][c]{#1}}
\caption{\textbf{3D gain by task.}
Accuracy (\%) with 2D and 3D inputs in each quartile of $A$, using the BERT~\cite{devlin-etal-2019-bert} backbone.
$\Delta$ and $n$ denote the 3D gain and the instance count, respectively.}
\label{tab:supp_quartile}
\footnotesize
\setlength{\tabcolsep}{3.5pt}
\renewcommand{\arraystretch}{1.1}
\begin{tabular*}{\columnwidth}{@{\extracolsep{\fill}}l l cccc c}
\toprule
Task & & $Q1$ & $Q2$ & $Q3$ & $Q4$ & All \\
\midrule
\multirow{3}{*}{STI}
 & 2D & 75.4 & 72.0 & 68.5 & 69.6 & 71.2 \\
 & 3D & 75.9 & 74.9 & 71.3 & 76.0 & 74.5 \\
 & $\Delta$ & \dpos{\dw{+0.5}} & \dpos{\dw{+2.9}} & \dpos{\dw{+2.8}} & \dpos{\dw{+6.4}} & \dpos{\dw{+3.3}} \\
\cmidrule(lr){2-7}
 & $n$ & 187 & 211 & 216 & 217 & 831 \\
\midrule
\multirow{3}{*}{PCR}
 & 2D & 65.4 & 60.6 & 72.0 & 60.4 & 64.6 \\
 & 3D & 66.5 & 63.2 & 76.7 & 62.4 & 67.1 \\
 & $\Delta$ & \dpos{\dw{+1.1}} & \dpos{\dw{+2.6}} & \dpos{\dw{+4.7}} & \dpos{\dw{+2.0}} & \dpos{\dw{+2.5}} \\
\cmidrule(lr){2-7}
 & $n$ & 179 & 155 & 150 & 149 & 633 \\
\bottomrule
\end{tabular*}
\vspace{-1mm}
\end{table}
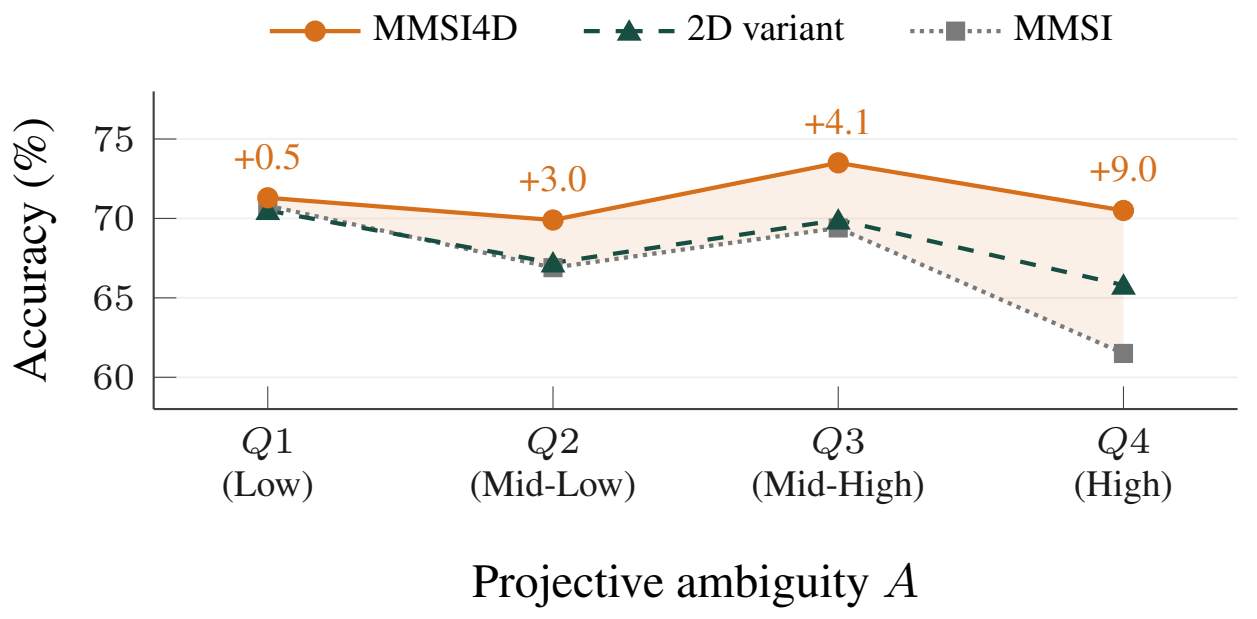
\begin{figure}[t]
\centering
\definecolor{cOurs}{HTML}{D66E1A}
\definecolor{cVar}{RGB}{22,78,66}
\definecolor{cPrior}{HTML}{7A7A7A}
\begin{tikzpicture}
\begin{axis}[
  width=1.05\columnwidth,  height=1.45in,
  xmin=0.6,
  xmax=4.4,
  ymin=58,
  ymax=78,
  xlabel={Projective ambiguity $A$},
  ylabel={Accuracy (\%)},
  xtick={1,2,3,4},
  xticklabels={{$Q1$\\ (Low)}, {$Q2$\\ (Mid-Low)}, {$Q3$\\ (Mid-High)}, {$Q4$\\ (High)}},
  ytick={60,65,70,75},
  ymajorgrids=true,
  grid style={gray!12, line width=0.35pt},
  axis x line*=bottom,
  axis y line*=left,
  axis line style={black!75},
  tick style={black!75},
  tick label style={font=\scriptsize, color=black!90},
  x tick label style={align=center, font=\scriptsize, color=black!90},
  label style={font=\small},
  xlabel style={at={(ticklabel cs:0.5)}, anchor=north, yshift=-1.0ex},
  legend style={
    at={(0.5,1.0)},
    anchor=south,
    yshift=5pt,
    inner sep=1pt,
    legend columns=3,
    draw=none,
    fill=none,
    font=\scriptsize,
    /tikz/every even column/.append style={column sep=1.0em},
  },
  clip=false,
  enlarge x limits=false,
  reverse legend,
]
\addplot[forget plot, draw=none, fill=cOurs, fill opacity=0.10]
  coordinates {(1,71.3) (2,69.9) (3,73.5) (4,70.5) (4,61.5) (3,69.4) (2,66.9) (1,70.8)} -- cycle;
\addplot[cPrior, line width=0.8pt, densely dotted, mark=square*, mark size=1.4pt, mark options={solid, fill=cPrior}]
  coordinates {(1,70.8) (2,66.9) (3,69.4) (4,61.5)};
\addlegendentry{MMSI}
\addplot[cVar, line width=0.8pt, dashed, mark=triangle*, mark size=1.9pt,
    mark options={solid, fill=cVar, yshift=-0.45pt},
    legend image code/.code={
      \draw[cVar, line width=0.8pt, dashed] (0cm,0cm) -- (0.3cm,0cm);
      \draw[cVar, line width=0.8pt, dashed] (0.6cm,0cm) -- (0.3cm,0cm);
      \draw[cVar, only marks, mark=triangle*, mark size=1.9pt,
        mark options={solid, fill=cVar, yshift=-0.45pt}] plot coordinates {(0.3cm,0cm)};
    }]
  coordinates {(1,70.5) (2,67.2) (3,69.9) (4,65.8)};
\addlegendentry{2D variant}
\addplot[cOurs, line width=0.8pt, mark=*, mark size=1.6pt, mark options={solid, fill=cOurs}]
  coordinates {(1,71.3) (2,69.9) (3,73.5) (4,70.5)};
\addlegendentry{MMSI4D}
\node[font=\scriptsize, text=cOurs, anchor=south] at (axis cs:1,71.9) {+0.5};
\node[font=\scriptsize, text=cOurs, anchor=south] at (axis cs:2,70.5) {+3.0};
\node[font=\scriptsize, text=cOurs, anchor=south] at (axis cs:3,74.1) {+4.1};
\node[font=\scriptsize, text=cOurs, anchor=south] at (axis cs:4,71.1) {+9.0};
\end{axis}
\end{tikzpicture}
\caption{\textbf{Comparison with MMSI across projective ambiguity.}
Accuracy (\%) in each quartile of $A$, using the BERT~\cite{devlin-etal-2019-bert} backbone.
Numbers denote the gain of MMSI4D over MMSI~\cite{lee2024modeling}.}
\label{fig:supp_quartile_mmsi}
\vspace{-2mm}
\end{figure}

\section{Additional Qualitative Results}
\phantomsection
\label{sec:supp_qual}

\new{Figures~\ref{fig:supp_qual_sti}--\ref{fig:supp_qual_q1} visualize the reconstructed 3D arrangement of four examples where MMSI4D corrects the prediction of its 2D variant.
The first two revisit the top examples of Fig.~3, and the other two are taken from opposite ends of the ambiguity range, $Q4$ and $Q1$.
Each figure illustrates the visual window with four frames sampled at roughly one-second intervals, together with the corresponding reconstruction.
In Figs.~\ref{fig:supp_qual_sti}--\ref{fig:supp_qual_q4}, the referent has a close neighbor on the image plane, making the two hard to disambiguate in 2D, whereas they are clearly separated in 3D.
Even in the low-ambiguity scene of Fig.~\ref{fig:supp_qual_q1}, where all candidates appear on the same side of the speaker, the reconstruction clearly separates the referent from the others.
These examples suggest that interpreting the speaker's cues together with the 3D arrangement of candidates resolves referents that remain ambiguous on the image plane.
In all four cases, MMSI4D successfully identifies the referent by contextualizing the visual cues with the dialogue.}

\begin{table}[t]
\centering
\caption{\textbf{Removing the 3D arrangement at test time.}
Average accuracy (\%) over both subsets with the BERT~\cite{devlin-etal-2019-bert} backbone.
Flattened denotes the same model evaluated with all participants placed at the same distance from the camera.}
\label{tab:flatten}
\footnotesize
\setlength{\tabcolsep}{5pt}
\renewcommand{\arraystretch}{1.15}
\resizebox{\linewidth}{!}{%
\begin{tabular}{l c >{\centering\arraybackslash}p{4.2em} >{\centering\arraybackslash}p{4.2em} cc}
\toprule
& Dimension & \multicolumn{2}{c}{Distance from Camera} & \multicolumn{2}{c}{Task} \\
\cmidrule(lr){2-2}\cmidrule(lr){3-4}\cmidrule(lr){5-6}
Model & 3D? & Training & Test & STI & PCR \\
\midrule
MMSI4D (Full)      & \cmark & \cmark & \cmark & 74.9 & 64.0 \\
MMSI4D (Flattened) & \cmark & \cmark & \xmark & \dneg{71.3} & \dneg{59.7} \\
\midrule
2D variant         & \xmark & \xmark & \xmark & 70.5 & 60.7 \\
\bottomrule
\end{tabular}}
\end{table}

\section{Limitations}
\phantomsection
\label{sec:supp_limitations}

\new{MMSI4D relies on several simplifying assumptions, each of which suggests a direction for future work.
First, it builds on monocular HMR with nominal intrinsics and a static camera, leaving room for methods that jointly reconstruct people and cameras from uncalibrated images or videos with moving cameras~\cite{muller2025reconstructing,chen2025human3r}.
Second, given that participants in the benchmark remain seated, we assume a stationary arrangement within a short visual window and summarize each candidate's relative position on the camera-$XZ$ plane with a temporal median.
However, people in everyday conversations often stand, walk, or regroup, which calls for modeling how the 3D arrangement evolves over time.
Third, we use head orientation as a proxy for attention, which could be captured in finer detail by integrating 3D-aware gaze estimation methods~\cite{fang2021dual,bao2022escnet,kawana2025ga3ce}.
More broadly, datasets of multi-party conversations that pair ground-truth 3D motion with fine-grained conversational annotations would allow models such as MMSI4D to be trained and evaluated at scale, free from reconstruction noise.}

\section{Broader Impacts}
\phantomsection
\label{sec:supp_impact}

\new{This work bridges 3D human reconstruction and multimodal social reasoning, showing that modeling multi-party conversations in 3D space helps identify whom each utterance addresses or mentions.
We believe this connection opens up opportunities for both future research and real-world applications, which we discuss below together with the considerations they raise.}

\new{On the research side, our results demonstrate that the spatial arrangement of participants, recovered from in-the-wild monocular video with an off-the-shelf HMR model, benefits social reasoning without requiring multi-camera capture.
This makes social interaction understanding a promising downstream task for HMR, where advances in reconstruction could bring further improvements.
The benefit of 3D inputs also holds across architectures and language backbones, suggesting that explicit per-participant 3D geometry could complement other models of social interaction.
This geometry could support related problems, including the prediction and generation of multi-party behavior.
Beyond the model itself, the projective ambiguity $A$ is computed from the reconstruction alone and can be applied to other conversational datasets to indicate where 3D reasoning is likely to matter, which may inform the design of future benchmarks.}

\new{On the application side, systems that assist or take part in group conversations need to know who is being addressed or mentioned at each moment.
Our approach, which takes a single monocular video as its visual input, could therefore serve as a building block for future embodied agents, meeting assistants, and assistive technologies for blind or low-vision users.
On the visual side, MMSI4D takes body keypoints and participant indices as input and does not rely on facial identity.
Nevertheless, it requires tracking each participant and associating them with their utterances, and its outputs describe the addressee or referent of each utterance.
This information can be sensitive, so systems built on our approach should be deployed with the consent of the people involved and appropriate privacy safeguards.
Finally, the way people use nonverbal cues such as head orientation and gestures may vary across cultures and individuals, which motivates evaluation across diverse populations.}

\newcommand{\suppqualexample}[1]{%
  \begin{tikzpicture}
    \node[anchor=south west, inner sep=0] (img) at (0,0)
      {\includegraphics[width=0.965\textwidth]{#1}};
    \begin{scope}[x={(img.south east)}, y={(img.north west)}]
      \node[rotate=90, font=\scriptsize, text=black!85] at (-0.014,0.842) {Input frames};
      \node[rotate=90, font=\scriptsize, text=black!85] at (-0.014,0.510) {Camera view};
      \node[rotate=90, font=\scriptsize, text=black!85] at (-0.014,0.168) {Top-down view};
    \end{scope}
  \end{tikzpicture}%
}

\begin{figure*}[p]
    \centering
    \suppqualexample{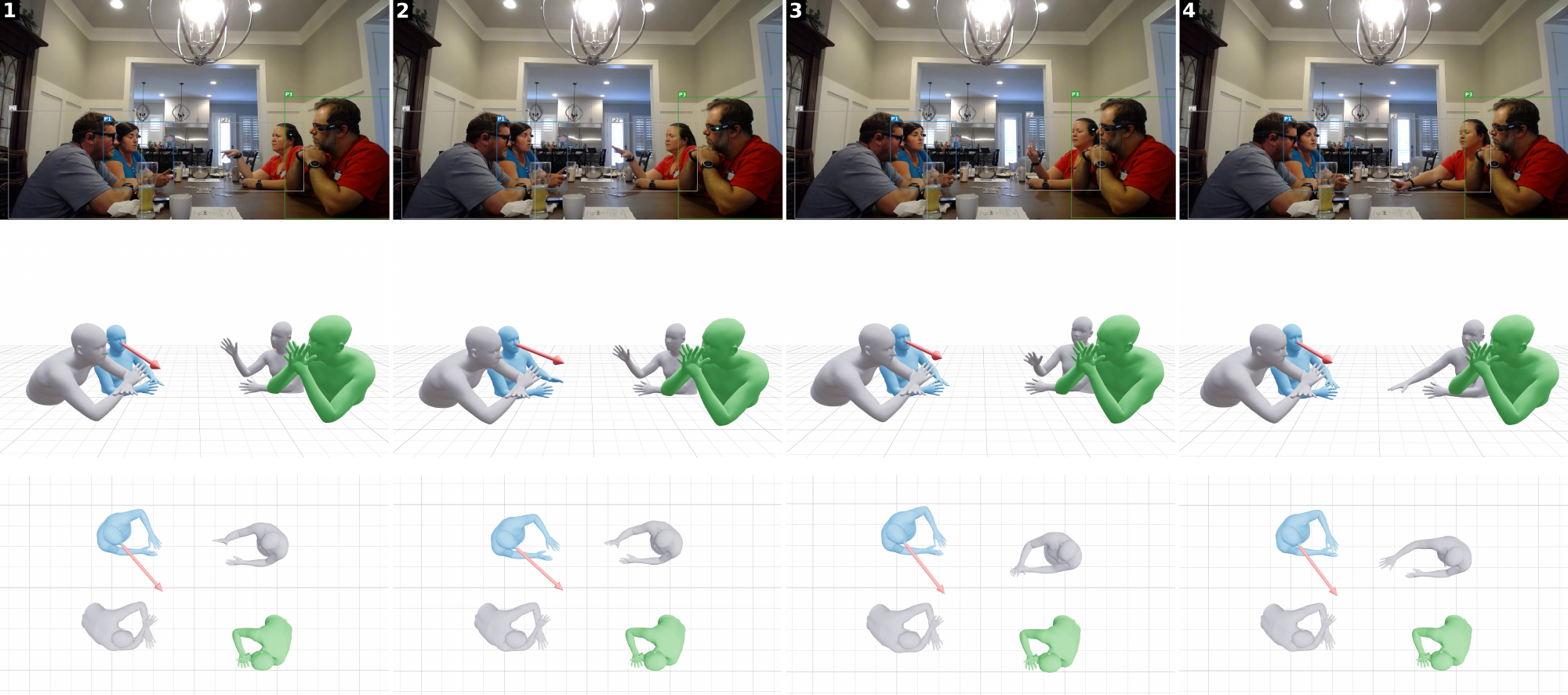}
    \caption{\textbf{Reconstructed 3D arrangement for the top-left example of Fig.~3.}
    From top to bottom, we show four input frames in temporal order, the reconstruction seen from the camera, and its top-down view with the camera at the bottom.
    The speaker is shown in \textcolor{blue}{blue}, with a \textcolor{red}{red} arrow for the head direction, and the ground-truth referent in \textcolor{green!70!black}{green}.
    The 2D variant predicts Player~2, who lies in nearly the same direction from the speaker on the image plane as the ground-truth referent (Player~3).
    In contrast, the two are clearly separated in 3D, as seen from above, and MMSI4D correctly identifies the referent.}
    \label{fig:supp_qual_sti}

    \vspace{3.0em}

    \suppqualexample{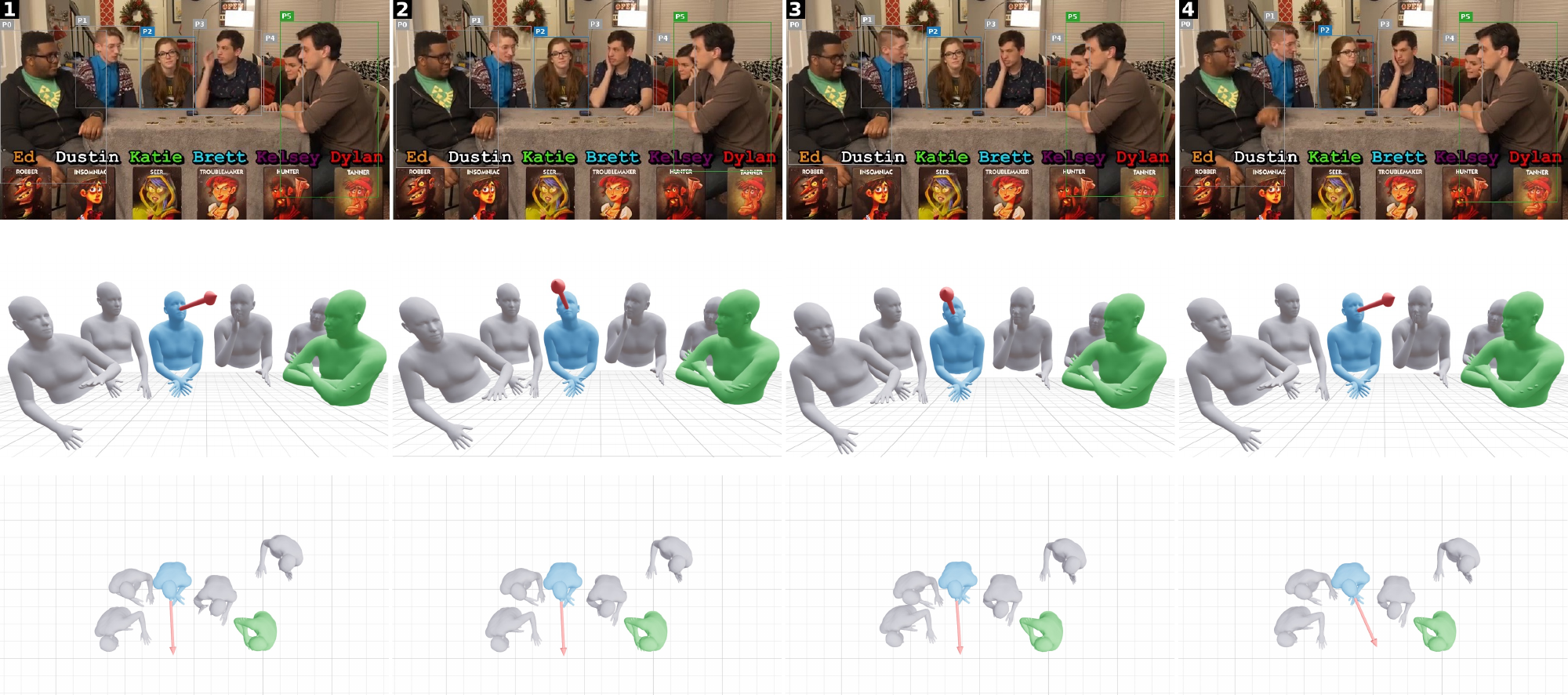}
    \caption{\textbf{Reconstructed 3D arrangement for the top-right example of Fig.~3.}
    Same format as Fig.~\ref{fig:supp_qual_sti}.
    The speaker faces the referent most directly in the final frame, a moment that the head--candidate compatibility captures by taking the maximum over frames.
    Despite some reconstruction noise from occlusion, the referent remains clearly distinguishable in the 3D arrangement.}
    \label{fig:supp_qual_pcr}
\end{figure*}

\begin{figure*}[p]
    \centering
    \suppqualexample{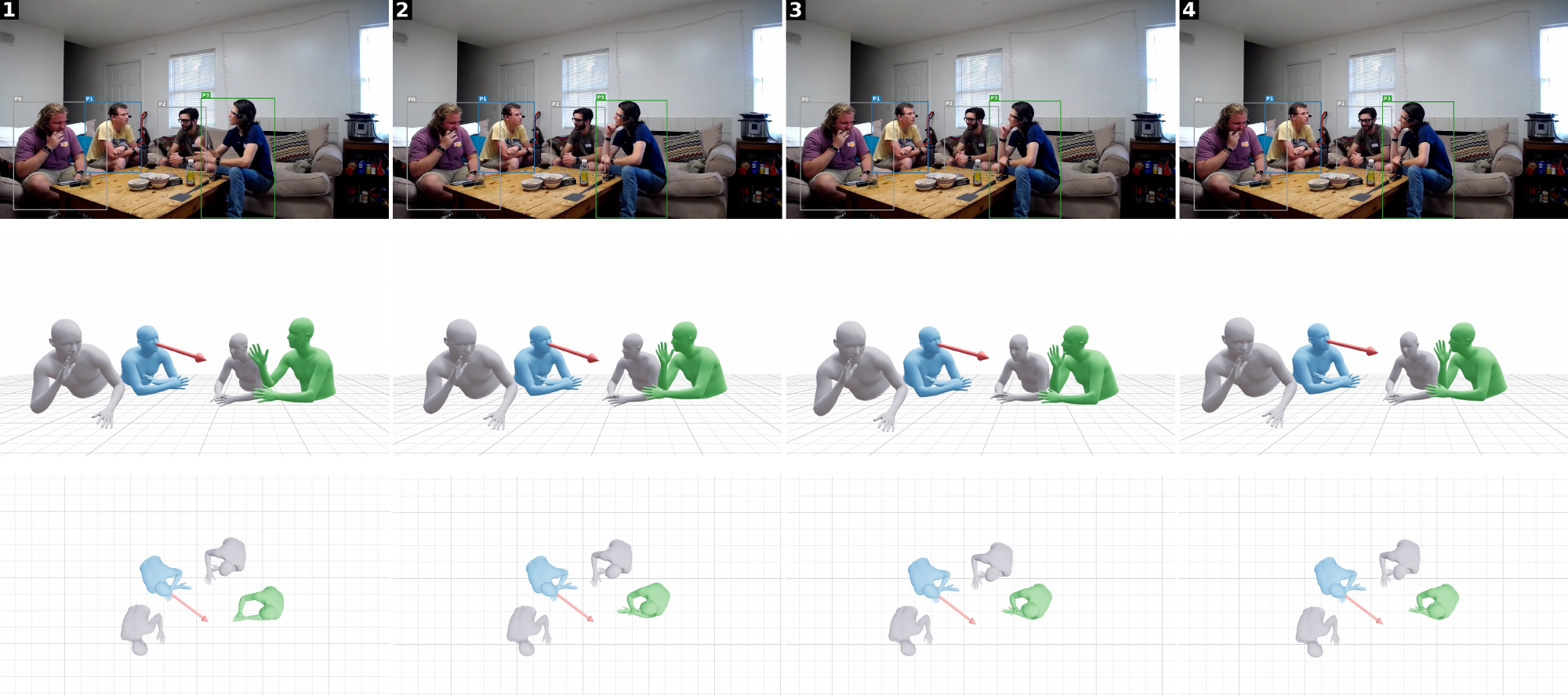}
    \caption{\textbf{Reconstructed 3D arrangement for a high-ambiguity example} in $Q4$ with $A=1.03$.
    Same format as Fig.~\ref{fig:supp_qual_sti}.
    The 2D variant predicts Player~2, an image-plane neighbor of the referent, whereas the two are well separated in 3D, with the head of the speaker oriented toward the referent.}
    \label{fig:supp_qual_q4}

    \vspace{3.0em}

    \suppqualexample{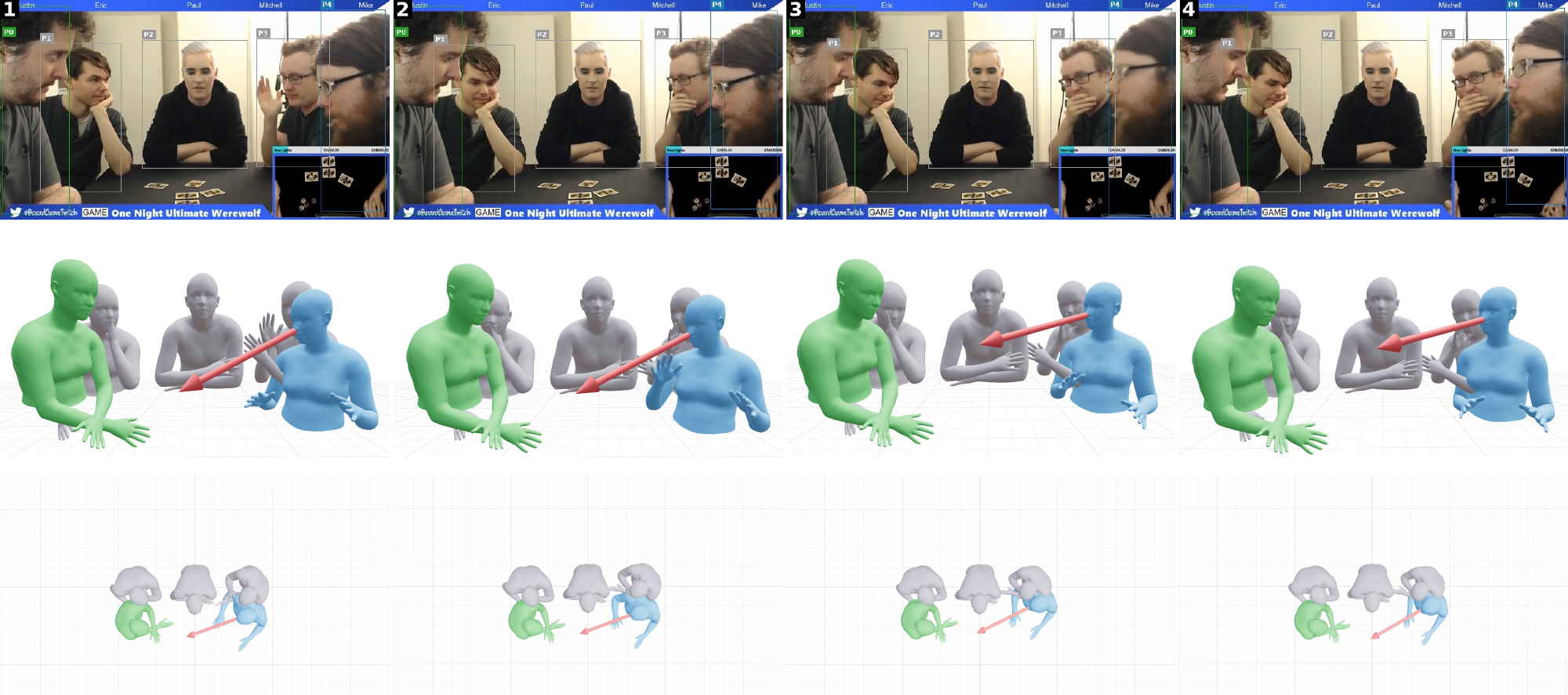}
    \caption{\textbf{Reconstructed 3D arrangement for a low-ambiguity example} in $Q1$ with $A=0.39$.
    Same format as Fig.~\ref{fig:supp_qual_sti}.
    All four candidates appear on the same side of the speaker in the image, and the 2D variant predicts Player~3.
    In 3D, the ground-truth referent (Player~0) sits slightly in front of the others, which allows MMSI4D to distinguish it.
    A picture-in-picture overlay partially occludes the body of the speaker, making its orientation difficult to recover.}
    \label{fig:supp_qual_q1}
\end{figure*}

\tcbset{
  lmmprompt/.style={
    colframe=gray!55,
    colback=gray!6,
    arc=0pt, outer arc=0pt,
    boxrule=0.5pt,
    left=6pt, right=6pt, top=6pt, bottom=6pt,
    width=\textwidth,
    fontupper=\footnotesize\ttfamily\raggedright\frenchspacing
  }
}

\begin{figure*}[p]
\centering
\begin{tcolorbox}[lmmprompt]
You are solving a multi-party social interaction benchmark.\\[0.5em]
Eight visual frames are shown in chronological order.\\[0.5em]
\includegraphics[width=\linewidth]{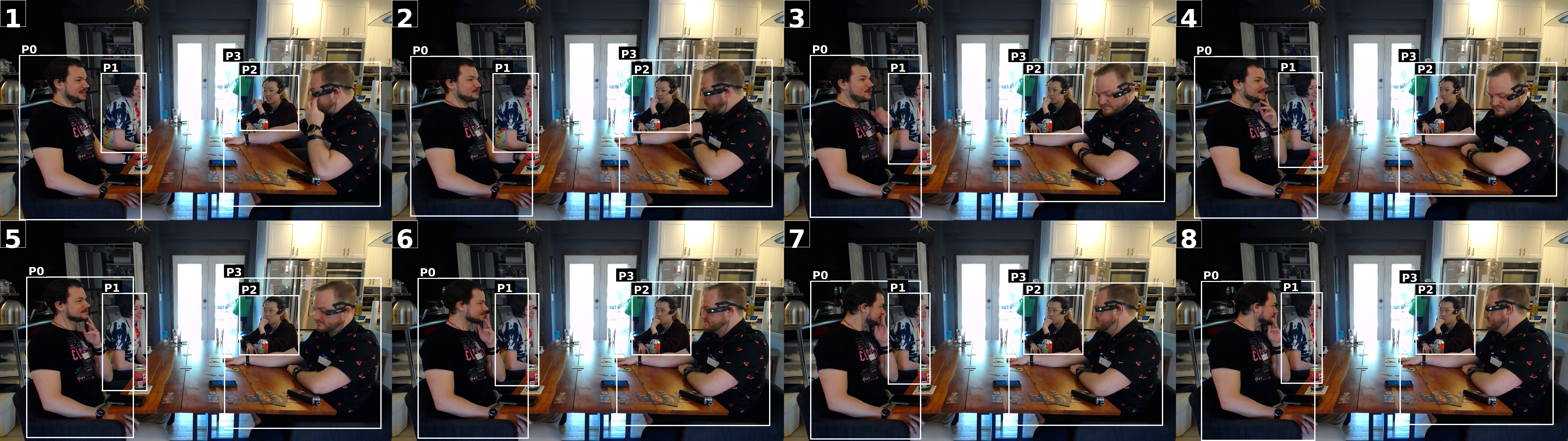}\\[0.5em]
Visible participants are labeled with participant IDs.\\
Use only the provided visual frames and conversation.\\[0.8em]
Task: Speaking Target Identification (STI)\\[0.8em]
Determine which participant the speaker of the TARGET utterance is directly speaking to.\\[0.8em]
Candidate IDs:\\
P0, P1, P2, P3\\[0.8em]
Conversation:\\
{[P1]}: But do we have to?\\
{[P3]}: We can all say we think it's in the center, and no one gets shot.\\
{[P1]}: Okay, well, it's either in the center or it's you. [crosstalk 00:04:19].\\
{[P3]}: It sucks. It's fifty-fifty. I've got to vote center, because I can't vote to shoot myself.\\
{[P0]}: You have no idea.\\
{[TARGET]} [P1]: Here's how it's going to go. We all vote to not shoot. If you are the Werewolf- \\
{[P3]}: Then I win. If I'm the Werewolf and you shoot center-\\
{[P2]}: So, we should shoot you just to be safe.\\
{[P1]}: No, no, no, we don't shoot center, we skip this round. If [crosstalk 00:04:40].\\
{[P2]}: It's only one round.\\
{[P3]}: It's one night Werewolf.\\[0.8em]
TARGET utterance:\\
{[P1]}: Here's how it's going to go. We all vote to not shoot. If you are the Werewolf-\\[0.8em]
Which participant is the speaker directly addressing?\\[0.8em]
Return exactly one candidate ID from the candidate list.\\
Do not provide an explanation.
\end{tcolorbox}
\caption{\textbf{Zero-shot video LMM input for STI (Ego4D).}
An example of the prompt in Sec.~\ref{sec:supp_prompt} filled with a test instance.
It contains the target utterance with its five preceding and five following utterances, together with eight input frames.
The prompt denotes each participant by a short ID, with Player~0 written as P0.
The ground-truth answer for this instance is P3.}
\label{fig:supp_lmm_input_sti}
\end{figure*}

\begin{figure*}[p]
\centering
\begin{tcolorbox}[lmmprompt]
You are solving a multi-party social interaction benchmark.\\[0.5em]
Eight visual frames are shown in chronological order.\\[0.5em]
\includegraphics[width=\linewidth]{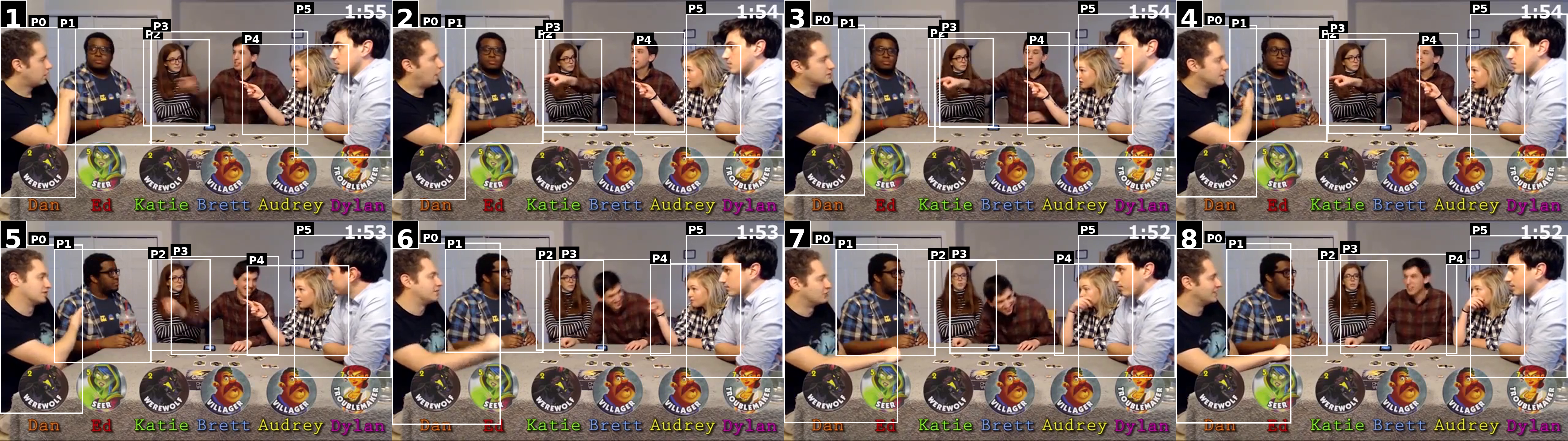}\\[0.5em]
Visible participants are labeled with participant IDs.\\
Use only the provided visual frames and conversation.\\[0.8em]
Task: Pronoun Coreference Resolution (PCR)\\[0.8em]
Determine which participant the marked pronoun in the TARGET utterance refers to.\\[0.8em]
Candidate IDs:\\
P0, P1, P2, P3, P4, P5\\[0.8em]
Conversation:\\
{[P4]}: Who's cards did you switch?\\
{[P1]}: 100\%.\\
{[P5]}: Them two, [P1] and [P0]'s.\\
{[P4]}: Okay so [P1] and [P0] switched, you were the Seer so that means you're the Seer so you could be a Werewolf if you have his card.\\
{[P3]}: But he wasn't- but he's-\\
{[TARGET]} [P0]: Yeah if [PRONOUN] was lying about being a Villager originally.\\
{[P2]}: Who was the robber?\\
{[P3]}: But he's saying he saw my Werewolf and I'm a Villager so I don't know why we trust him.\\
{[P2]}: You were the Troublemaker, who was the Robber?\\
{[P4]}: Because I trust [P0] implicitly.\\
{[P0]}: Thank you.\\[0.8em]
TARGET utterance:\\
{[P0]}: Yeah if [PRONOUN] was lying about being a Villager originally.\\[0.8em]
Which participant does the marked pronoun refer to?\\[0.8em]
Return exactly one candidate ID from the candidate list.\\
Do not provide an explanation.
\end{tcolorbox}
\caption{\textbf{Zero-shot video LMM input for PCR (YouTube).}
Same format as Fig.~\ref{fig:supp_lmm_input_sti}, where the pronoun to be resolved is marked with \mbox{\texttt{[PRONOUN]}}.
The ground-truth answer for this instance is P1.}
\label{fig:supp_lmm_input_pcr}
\end{figure*}

\end{document}